\documentclass[lettersize,journal]{IEEEtran}
\usepackage{amsmath,amsfonts}
\usepackage{algorithmic}
\usepackage{array}
\usepackage[caption=false,font=normalsize,labelfont=sf,textfont=sf]{subfig}
\usepackage{textcomp}
\usepackage{stfloats}
\usepackage{url}
\usepackage{verbatim}
\usepackage{graphicx}
\def\BibTeX{{\rm B\kern-.05em{\sc i\kern-.025em b}\kern-.08em
    T\kern-.1667em\lower.7ex\hbox{E}\kern-.125emX}}
\usepackage{balance}
\usepackage{booktabs}
\usepackage{multirow}
\usepackage{multicol}
\usepackage{color,xcolor}
\usepackage{amssymb}
\usepackage[ruled, vlined]{algorithm2e}

\newcommand*{\red}{\textcolor{black}}

\begin{document}

\title{Cross-Modal Causal Representation Learning for Radiology Report Generation}

\author{Weixing~Chen, Yang~Liu,~\IEEEmembership{Member,~IEEE}, Ce~Wang, Jiarui~Zhu, Guanbin~Li,~\IEEEmembership{Member,~IEEE}, \\Cheng-Lin Liu,~\IEEEmembership{Fellow,~IEEE}, and~Liang~Lin,~\IEEEmembership{Fellow,~IEEE}
\thanks{This work is supported in part by the National Key R\&D Program of China under Grant No.2021ZD0111601, in part by the National Natural Science Foundation of China under Grant No.62436009, No. 62322608, and No.62301532, in part by the Guangdong Basic and Applied Basic Research Foundation under Grant No.2025A1515011874 and No.2023A1515011530. (\emph{Corresponding author: Yang Liu.})} 
\thanks{Weixing Chen, Yang Liu, Guanbin Li and Liang Lin are with the School
of Computer Science and Engineering, Sun Yat-sen University, China, and Guangdong Key Laboratory of Big Data Analysis and Processing, Guangzhou, China. \protect
(E-mail: chenwx228@mail2.sysu.edu.cn, liuy856@mail.sysu.edu.cn, liguanbin@mail.sysu.edu.cn, linliang@ieee.org)}
\thanks{Ce Wang is with the School of Science, Sun Yat-sen University, Shenzhen, 510275, China. (E-mail: fogever@icloud.com)}
\thanks{Jiarui Zhu is with the Hong Kong Polytechnic University. (E-mail: jiarui.zhu@connect.polyu.hk)}
\thanks{Cheng-Lin Liu is with the State Key Laboratory of Multimodal Artificial Intelligence Systems, Institute of Automation, Chinese Academy of Sciences, Beijing, China, and also with the School of Artificial Intelligence, University of Chinese Academy of Sciences, Beijing, China. (liucl@nlpr.ia.ac.cn)}
}


\maketitle

\begin{abstract}
 \red{Radiology Report Generation (RRG) is essential for computer-aided diagnosis and medication guidance, which can relieve the heavy burden of radiologists by automatically generating the corresponding radiology reports according to the given radiology image.
 However, generating accurate lesion descriptions remains challenging due to spurious correlations from visual-linguistic biases and inherent limitations of radiological imaging, such as low resolution and noise interference.
 To address these issues, we propose a two-stage framework named Cross-Modal Causal Representation Learning (CMCRL), consisting of the Radiological Cross-modal Alignment and Reconstruction Enhanced (RadCARE) pre-training and the Visual-Linguistic Causal Intervention (VLCI) fine-tuning.
 In the pre-training stage, RadCARE introduces a degradation-aware masked image restoration strategy tailored for radiological images, which reconstructs high-resolution patches from low-resolution inputs to mitigate noise and detail loss. Combined with a multiway architecture and four adaptive training strategies (e.g., text postfix generation with degraded images and text prefixes), RadCARE establishes robust cross-modal correlations even with incomplete data.  
 In the \red{VLCI} phase, we deploy causal front-door intervention through two modules: the Visual Deconfounding Module (VDM) disentangles local-global features without fine-grained annotations, while the Linguistic Deconfounding Module (LDM) eliminates context bias without external terminology databases. Experiments on IU-Xray and MIMIC-CXR show that our CMCRL pipeline significantly outperforms state-of-the-art methods, with ablation studies confirming the necessity of both stages. Code and models are available at \url{https://github.com/WissingChen/CMCRL}.}
\end{abstract}

\begin{IEEEkeywords}
Radiology Report Generation, causality, visual-language pre-training.
\end{IEEEkeywords}

\section{Introduction}

\begin{figure}[t]
 \centering
    \includegraphics[width=1\linewidth]{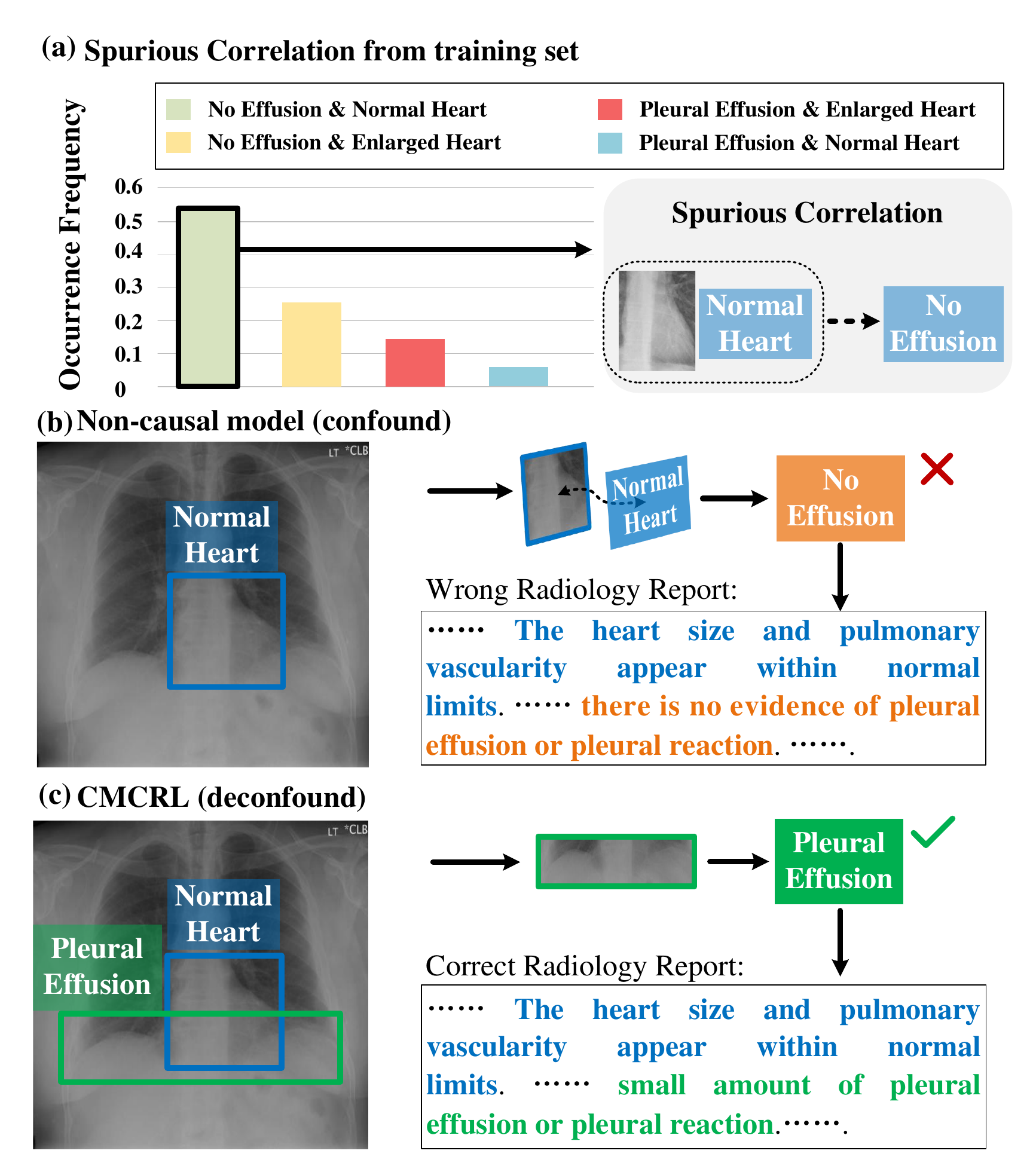}
    \caption{The example of visual-linguistic spurious correlation of {RRG}{. (a) There exists data bias in the training set, where the combination of ``No Effusion" and ``Normal Heart" occurs significantly more frequently than others. This leads the model to correlate the ``Normal Heart" with the conclusion of ``No Effusion", forming a spurious correlation. (b) Consequently, the non-causal model does not truly rely on pulmonary information to determine the presence of effusion. Instead, it arrives at the conclusion of ``No Effusion" based on the presence of a ``Normal Heart". (c) In contrast, our \red{CMCRL}, by extracting critical visual features and applying them to causal front-door intervention, enables the model to identify features that are causally related to ``Pleural Effusion" and produce the correct report.}
    }
\label{fig:intro_bias_info}
\end{figure}

\IEEEPARstart{R}{adiology} images are widely used in clinical procedures \cite{yu2022crosslink}, providing significant evidence for disease analysis and radiological diagnosis~\cite{zhou2021review}. Nevertheless, observing suspicious lesions and writing a coherent diagnosis report is time-consuming, even for experienced radiologists. Furthermore, inexperienced radiologists often fail to capture tiny abnormalities due to the high requirement for clinical knowledge. To relieve this issue, Radiology Report Generation (RRG) has emerged and attracted growing interest in recent years~\cite {tanida2023interactive}. RRG extracts features from radiology images and generates the corresponding reports, which is similar to image captioning~\cite{nguyen2022effective}. However, the current RRG faces three challenges that are significantly different from the image captioning task:
\textbf{1)} Longer sentence generation (60-100 tokens) tends to accumulate a larger bias, whereas image captions typically consist of fewer than 20 tokens~\cite{chen2020generating}, 
\textbf{2)} The necessity to capture all key regions in the radiology image (i.e., abnormalities and diagnostic evidence) and a low tolerance for factually incorrect information~\cite{tanida2023interactive}, 
\textbf{3)} More complex linguistic and visual semantic patterns require a proficient understanding of radiological information, whereas entities in natural images are diverse and easily distinguishable~\cite{zhang2020radiology, zhou2021review}. 
Therefore, these challenges impose significant limitations on modeling visual-linguistic interactions and learning informative cross-modal representations for accurate RRG~\cite{chen2020generating,chen2023cross}.

To tackle the aforementioned challenges, current RRG methods have made significant efforts, such as the memory-driven module for longer sentence generation~\cite{chen2020generating}, additional knowledge for more accurate description~\cite{liu2021exploring}, and contrast with normal samples (i.e., images without lesion areas) for the capture of abnormalities (i.e., lesion areas within images)~\cite{liu2021contrastive}. Most of the previous RRG methods aim to capture the latent subtle differences in images (visual biases) and learn a concise set of key descriptions in the text (linguistic biases) for accurate long-sequence generation. However, these methods usually focus on training computationally expensive models based on a large number of region-level annotations~\cite{tanida2023interactive} and task-specific knowledge\footnote{These methods build the template or knowledge database laboriously, making it hard to transfer those approaches directly to other datasets~\cite{yang2022knowledge}.}, rather than focusing on identifying and mitigating biases inherent in the data itself.

As shown in Fig \ref{fig:intro_bias_info} (a-b), the significant visual and linguistic biases in numerous image-text data build a spurious correlation between ``Normal Heart" and ``No Effusion", leading to the incorrect and unreliable reports. To mitigate the cross-modal bias and encourage the model to infer based on the corresponding visual basis, as shown in Fig \ref{fig:intro_bias_info} (c), causal inference \cite{pearl2016causal} has shown promising performance in several visual-linguistic tasks \cite{liu2022causal}. However, directly applying existing causal methods to the RRG task may yield unsatisfactory results due to the unobservable bias in the visual and linguistic domain and the complex visual-linguistic interaction in radiology images and textual reports. Although back-door intervention can cut off the shortcut path, it requires approximating the observable bias using a well-trained visual object detector or a well-constructed linguistic dictionary. Fortunately, causal front-door intervention gives a feasible way to mitigate visual-linguistic spurious correlations without the calculation of unobservable bias. With causal front-door intervention, we can eliminate the spurious cross-modal correlations effectively by introducing an additional mediator~\cite{liu2023cross, yang2021causal} and generate an accurate description of ``Pleural Effusion". The mediator can be assumed as the feature of ``Pleural" in different findings (e.g., ``Normal Heart") to estimate the sub-distribution of ``Pleural Effusion"~\cite{yang2021deconfounded}. However, the accurate and reasonable acquisition of the mediator is challenging, especially without the support of additional radiological knowledge. 

\red{To effectively obtain reliable estimates of mediator, well-aligned radiological multi-modal representations are crucial. However, existing multi-modal approaches are limited by singular pre-training strategies and fail to fully leverage data with missing modalities within datasets~\cite{liu2023m3ae}. Moreover, given the issues of low-resolution inputs and noise interference present in radiological data, directly adopting multi-modal pre-training models designed for natural image tasks does not effectively capture and align multi-modal features~\cite{he2022masked, wang2021simvlm}. This further restricts the performance of subsequent causal interventions.}

\red{Motivated by the characteristic multi-modal radiological data and effectiveness of causal inference in deconfounding the cross-modal bias, we introduce a Cross-Modal Causal Representation Learning (CMCRL) framework for RRG. This framework includes a pre-training model specifically designed for radiological data, termed Radiological Cross-modal Alignment and Reconstruction Enhanced (RadCARE), and a lightweight cross-modal causal intervention model without requiring an observable bias assumption, named Visual-Linguistic Causal Intervention (VLCI), to mitigate biases in visual and linguistic data.
Our RadCARE framework introduces a degradation-aware masked image restoration strategy that reconstructs high-resolution patches from degraded inputs, explicitly mitigating detail loss while preserving anatomical coherence. This is coupled with a multi-way architecture enabling four adaptive training strategies, including postfix text generation via prefix text or multi-modal data, and masked image restoration via degraded images or combined with complete text.
Our VLCI framework contains the visual deconfounding module (\textbf{VDM}) and linguistic deconfounding module (\textbf{LDM}) based on the causal front-door intervention paradigm. In VDM, the visual mediator is constructed by local detail information (e.g., lung texture) and global contour (e.g., lung contour) from radiology images, to disentangle the visual features. The linguistic confounders can be eliminated by the LDM, which estimates the change in the probability of word embedding caused by visual details and linguistic context. 
In summary, our main contributions are listed as follows}:

\begin{itemize}
    \item \red{To alleviate the problem of unpaired data and capture detailed features when pre-training cross-modal data, we propose RadCARE that integrates degradation-aware masked image restoration with postfix text generation for pre-training in various data situations (e.g., unpaired, single modality), which is efficient and easy to implement.}
    \item  \red{To mitigate cross-modal biases, we propose visual-linguistic causal intervention modules VDM and LDM, integrated into the VLCI without requiring additional data or a well-trained semantic extractor for guidance. 
    } 
    \item \red{We propose a CMCRL framework for RRG, which introduces mediators without additional knowledge, to deconfound the visual-linguistic features by causal front-door intervention. To the best of our knowledge, we are the first to conduct a cross-modal causal intervention for RRG. Experiments show that we achieve state-of-the-art performance on IU-Xray and MIMIC-CXR datasets.}
\end{itemize}

\section{Related Work}

\subsection{Image Captioning}

Image captioning, based on the encoder-decoder framework, aims to describe image content in text. Recent advancements have enhanced performance in three areas: visual representation learning, linguistic generation, and training strategies~\cite{stefanini2022show}. While CNN-based global feature extraction compresses information excessively~\cite{karpathy2015deep}, integrating visual saliency through regional features and combining visual-semantic data improves results~\cite{jiang2022visual}. In linguistic processing, coarse-to-fine captioning~\cite{wang2017skeleton} and multi-layer LSTM integration~\cite{xian2019self} have shown promise but are limited by training efficiency. Transformer-based models~\cite{devlin2018bert} leverage self-attention for modality learning and cross-attention for multi-modal integration, significantly boosting performance~\cite{nguyen2022grit, liu2022show}. Performance is further improved by training strategies such as sample order optimization~\cite{zhou2019re}, reinforcement learning~\cite{nguyen2022effective}, and visual-linguistic pre-training~\cite{yu2022coca, wang2021simvlm}. 

\red{With the proliferation of Large Language Models (LLMs), image captioning achieved significant breakthroughs in both model architecture and generation capabilities. The V2L-Tokenizer~\cite{zhu2024beyond} model demonstrated SOTA performance across multiple tasks through a non-finetuning approach. Meanwhile, prompt learning offered new insights into the controllability of image captioning~\cite{liu2024improved, chen2024sharegpt4v}. Additionally, models like GPT-4o\cite{hurst2024gpt} and Qwen2-VL\cite{bai2023qwen} integrated advanced visual processing capabilities with powerful language generation, achieving remarkable results without extensive fine-tuning.} 
Compared to image captioning, RRG shares similar structures~\cite{stefanini2022show} but focuses on subtle lesion detection in radiology images, generating longer and more complex descriptions.

\subsection{Causal Inference}
Causality offers a robust approach to mitigating the visual-linguistic bias stemming from the heterogeneity of multi-modal data by eliminating spurious correlations~\cite{pearl2016causal,liu2022causal}. Causal inference~\cite{liu2022causal} estimates hidden causal effects within distributions, significantly improving model generalization by addressing confounders through back-door, front-door, or counterfactual interventions. This method enhances performance in tasks like image classification~\cite{yue2020interventional}, semantic segmentation~\cite{miao2023caussl}, visual feature representation~\cite{liu2022contextual, wang2021causal}, image captioning~\cite{liu2022show}, and visual question answering (VQA)\cite{liu2023cross}. For instance, Wang et al.\cite{wang2020visual} applied back-door intervention to improve Faster R-CNN, which subsequently enhanced performance in VQA~\cite{zang2023discovering} and image captioning~\cite{liu2022show}. Additionally, Liu et al. proposed an event-level causal VQA model using front-door intervention, which leverages attention to integrate local and global causality-aware visual-linguistic representations~\cite{liu2023cross}. This approach also addresses confounding by simulating causal interventions based on human priors~\cite{yang2023good}.

Since confounders are often unobservable, front-door and counterfactual interventions are particularly effective. These interventions can capture hidden confounders by integrating causal mechanisms into attention modules through cross-sample attention or self-annotation~\cite{hu2021causal, wang2021causal, xue2023variational}. 
\red{In the era of LLMs, causal learning has emerged as a key direction for enhancing both the generalization ability and interpretability. On one hand, by integrating causality into the processes of pre-training, fine-tuning, and inference, LLMs can better handle complex tasks, while also improving their performance in multi-modal scenarios~\cite{zhao2024causal, chu2024causal}. On the other hand, causal learning enables LLMs to move beyond mere statistical correlations and understand the underlying causal mechanisms in the data, thereby generating more reliable and ethically aligned outputs~\cite{huang2025causality}.}
Unlike prior work focusing on VQA or image captioning, we tackle the more complex task of RRG, which involves modeling intricate visual-linguistic interactions in radiology images and textual reports~\cite{zhao2018causaltriad, miao2023caussl, li2023causally}. Our proposed front-door causal intervention method addresses and eliminates both visual and linguistic spurious correlations.


\subsection{Radiology Report Generation}
Recently, RRG methods, inspired by image captioning, have achieved notable success. However, abnormal descriptions and lesion regions make up only a small portion of patient samples, leading to visual-linguistic bias. Knowledge-aware approaches address this by leveraging external knowledge to identify abnormalities and describe them accurately~\cite{zhang2020radiology, liu2021exploring}. Although fixed knowledge is limited for complex radiological cases, dynamic knowledge construction using RadGraph~\cite{jain2021radgraph} has been proposed~\cite{yang2022knowledge, li2023dynamic}. Furthermore, external knowledge can be integrated through label extractors or manually annotated regions to enhance report generation~\cite{you2021aligntransformer, wang2022cross, tanida2023interactive}. 

While knowledge-based models show promise, acquiring knowledge is costly and difficult to generalize. To mitigate this, CA~\cite{liu2021contrastive}, CMCL~\cite{liu2022competence}, CAMANet~\cite{wang2024camanet} explored abnormal regions by comparing them with normal samples. Templates based on similar images also help reduce representation distortion~\cite{yang2022knowledge, li2023dynamic}. Memory-driven transformers can store visual~\cite{nooralahzadeh2021progressive,divya2024memory} and linguistic~\cite{chen2020generating} contexts without templates but risk cross-modal bias. To address this, models align cross-modal data in embedding spaces for retrieval~\cite{chen2022cross, wang2022cross}. 
\red{With the advancement of multi-modal large language models (MLLMs), their robust visual reasoning capabilities, internal knowledge, and language generation skills have significantly enhanced performance in radiology report generation tasks. However, the inherent tendency of large models to produce hallucinations continues to pose challenges for this task. Consequently, some methods have focused on developing approaches that enable large models to generate descriptions consistent with the findings in radiological images~\cite{liu2024context, jin2024promptmrg, wang2023r2gengpt}. } 
To further reduce cross-modal bias, we introduce a lightweight VLCI approach, which applies causal front-door intervention to uncover true cross-modal causality and reduce reliance on additional annotations.

\section{Methodology} 

In this section, we first introduce the RadCARE framework for pre-training, followed by the two essential cross-modal causal intervention modules, i.e., the Visual Deconfounding Module (VDM) and the Linguistic Deconfounding Module (LDM). Then, we introduce how to integrate VDM and LDM into the VLCI for cross-modal causal intervention and complete the CMCRL framework.

\subsection{Overview}

\begin{figure}
    \centering
    \includegraphics[width=1\linewidth]{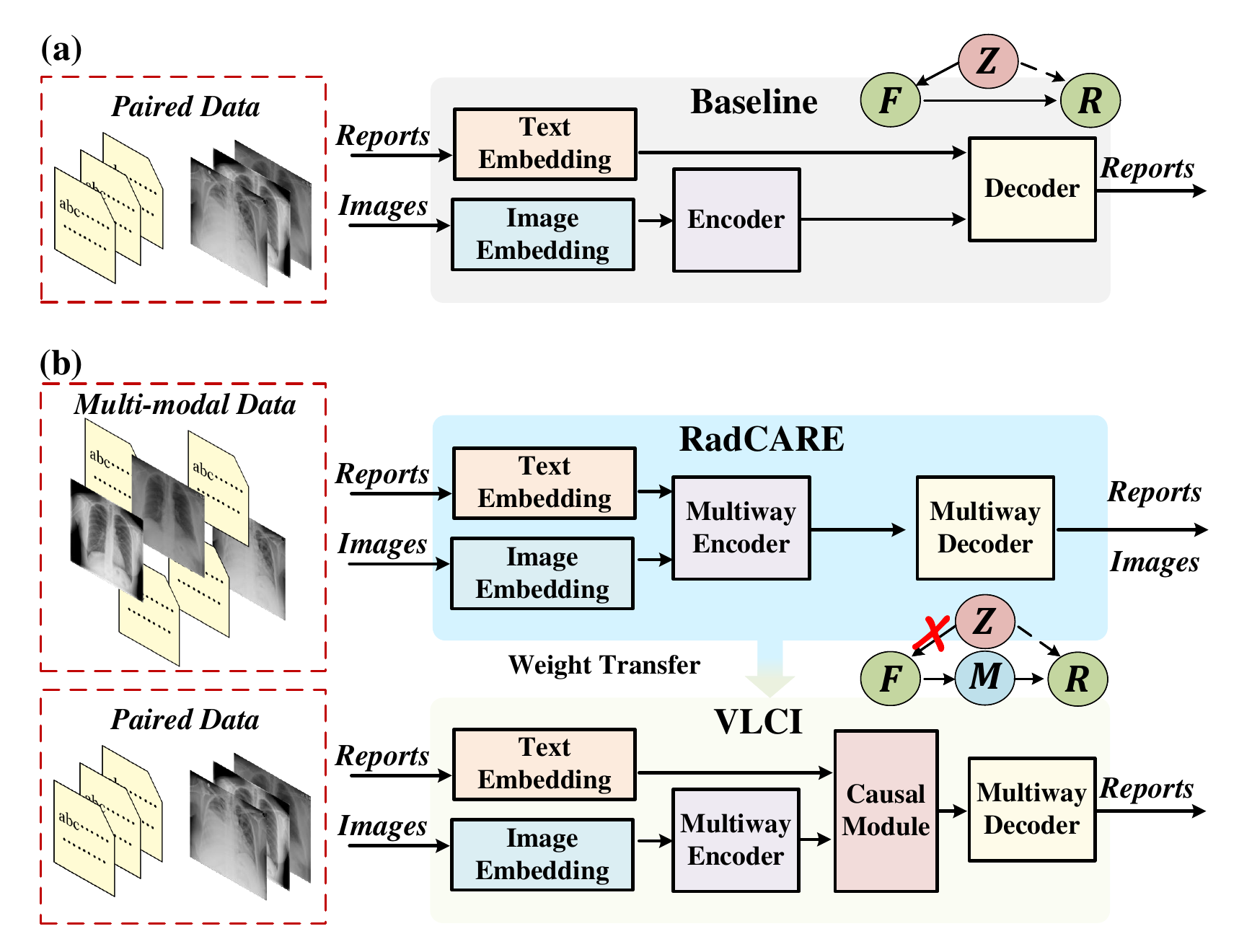}
\caption{\red{Comparison of the pipeline for (a) Baseline, i.e., non-causal model, and (b) our CMCRL framework. 
    }}
        \label{fig:method_overview}
\end{figure}

\red{
A typical RRG model can generate a series of radiological findings and insights given a radiological image. Upon receiving the Chest X-ray image $I\in\mathbb{R}^{C\times H\times W}$, the model first utilizes a visual feature extractor to obtain $h_v$, which guides the generation of the next word $w_i \in R$ from the prefix text embedding $h_w$, as shown in Fig.~\ref{fig:method_overview} (a). 
However, due to the presence of confounders $Z$ in both the visual and linguistic modalities, the non-causal model may capture the spurious correlations between ``Normal Heart" and ``No Effusion", which causes the neglect of ``Pleural Effusion" accompanied by ``Normal Heart", as shown in Fig.~\ref{fig:method_overview} (c). 
Fortunately, the mediators $M$ can cut off the link between $Z$ and $F$, and $F=\{h_v,h_w\}$ is denoted the multi-modal feature. Mediators enable the intervention and adjustment of the relation between $F$ and $R$, thereby revealing the true causal relation between the two variables. 
Thus, we perform a causal intervention on the multi-modal feature prior to the decoder to estimate the deconfounded probability of the correct word, as shown in Fig.~\ref{fig:method_overview} (b). 
Nevertheless, due to the absence of elaborate datasets and a well-trained feature extractor, we conduct a causal front-door intervention to eliminate the spurious correlations via mediators $M$. However, the estimation of both confounders and the mediator requires sufficient prior information, i.e., various visual-linguistic concepts. Therefore, we leverage the RadCARE to construct the correlation between the visual contexts and linguistic concepts before conducting the causal intervention using VLCI.
}

\subsection{RadCARE}

In the radiological pre-training framework, there exist two difficulties: (1) the unpaired data that only has a single modality is hard to be utilized in supervised learning, and (2) heterogeneous data that makes it difficult to distinguish the region features because the morphology of the same lesion varies greatly~\cite{zhou2021review}. 
To address these challenges and provide fine-grained region features without region labels, we employ a visual-linguistic pre-training approach to learn and align visual-linguistic information, as shown in Fig.~\ref{fig:method_overview} (b). 
\red{
Unlike existing methods that simply combine Masked Language Modeling (MLM) with Masked Image Modeling (MIM) \cite{zhao2023mamo} or rely on contrastive learning for cross-modal alignment \cite{xvlm,singh2022flava,varma2023villa}, 
we integrate postfix text generation task with degradation-aware masked image restoration task to achieve efficient cross-modal representation learning and alignment. 
To address the issue of missing modalities, we implement alternating training using multiple data input schemes, surpassing existing approaches that rely on a single training strategy. 
Moreover, to effectively capture the subtle details in radiological images, we employ a degradation-aware mechanism for pixel-level cross-modal semantic learning, thereby enhancing the alignment of visual and linguistic information.
}

\begin{figure}
    \centering
    \includegraphics[width=1\linewidth]{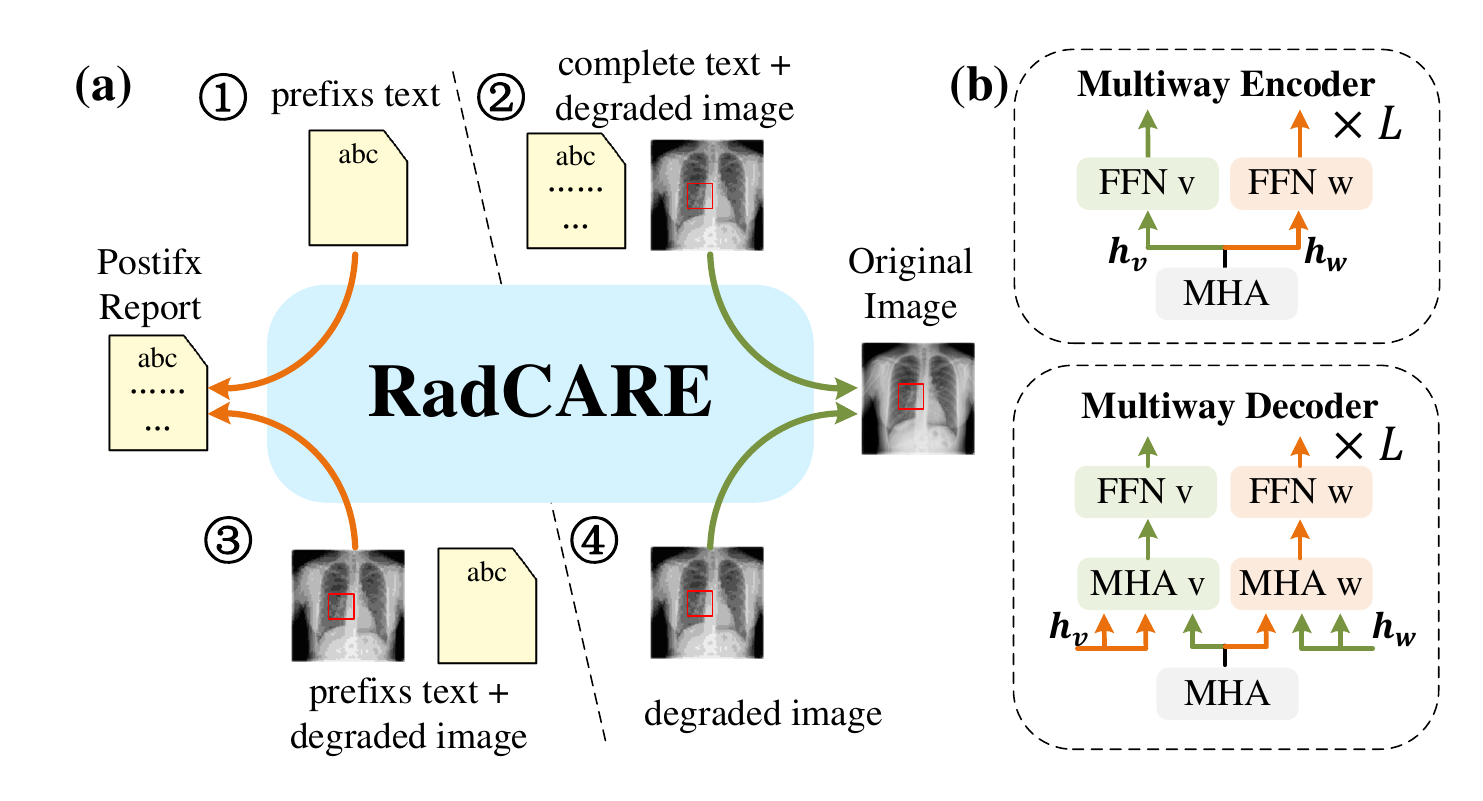}   
    \caption{\red{(a) shows four training strategies of our RadCARE, (b) demonstrates the detail of the multiway transformer blocks.}}
        \label{fig:method_radcare}
\end{figure}



\red{Specifically, RadCARE introduces four strategies for pre-training as shown in Fig~\ref{fig:method_radcare} (a), including 
1) generation of postfix text with text prefixes as input, 
2) restoration of the masked image with both complete text and degraded images as input,
3) generation of postfix text with degraded images and text prefixes as input, and
4) restoration of the masked image with degraded images as input.
}
To enable flexible switching between various strategies, we use a multiway transformer to extract multi-modal features and two linear layers to solve \red{text generation and image restoration tasks}, respectively~\cite{wang2021simvlm, he2022masked}, as shown in Fig\ref{fig:method_overview} (b). In each block of the multiway encoder, the attention layer is weight-shared while the two feed-forward layers handle the corresponding modal features respectively~\cite{wang2022image}. Similarly, each block of the multiway decoder consists of a weight-shared self-attention layer, a pool of feed-forward networks used for different modalities, as shown in Fig.~\ref{fig:method_radcare} (b). Additionally, the noise present in the MIMIC-CXR dataset, such as tasks requiring differentiation between the left and right lung despite being given lateral views, hinders the model's ability to learn robustly \cite{tanno2019learning, wang2024robust}. To address this issue, we combined two datasets for pre-training, as the IU-Xray dataset contains both lateral and frontal views.

\subsubsection{\red{Postfix Text Generation}} Motivated by the work of SimVLM~\cite{wang2021simvlm}, we extract image features from the first three blocks of ResNet101~\cite{he2016deep} as prefix tokens for the \red{RadCARE}. Simultaneously, the text is randomly divided into two parts, with one part generated by another under the guidance of the obtained image tokens. \red{In cases where the corresponding image is missing, the \red{RadCARE} can still be trained using only the text.} Let $h_v \in \mathbb{R}^{\frac{HW}{P^2}\times d}$ denote the image token extracted from the raw image $I$, where $P$ represents the patch size and $d$ is the embedding size. Then, $\{{w}_{{np}}, \ldots, {w}_{n}\}$ represents the postfix sequence following the textual description $h_w$ of length $n_p\ge 0$. Thus, the formulation is as follows:
\begin{equation}
    \mathcal{L}_{\textrm{\red{text}}}(\theta) = -\sum^{n}_{i=n_p}logP_\theta(w_{i}|h_v, h_{w_{<n_p}}),
\end{equation}
Here, $\theta$ denotes the trainable parameters of the model, $h_v$ represents the visual embedding with a trainable 2D positional encoding, $h_w$ is learned based on a fixed vocabulary and serves as the prefix received by the encoder, and $n$ denotes the length of the report.


\subsubsection{\red{Masked Image Restoration}} {To handle unpaired images, we leverage the MIM paradigm~\cite{he2022masked}. 
Furthermore, since \red{masked image restoration} can be trained using pairwise data~\cite{geng2022multimodal}, the missing semantics of masked images can be complemented by text, thereby enhancing cross-modal association. Additionally, we learn radiological image representations by reconstructing high-resolution patches from low-resolution inputs, which can encode more local information into latent embeddings~\cite{zhou2023advancing}.
Consequently, we sub-sample the images for degradation before the visual embedding and reconstruct the masked visual token extracted from CNNs by incorporating the semantics of both the unmasked visual token and the linguistic token.
This degradation-aware approach enables us to capture subtle differences in the dataset~\cite{huang2021gloria,cheng2023prior}.} The objective of \red{masked image restoration} can be formulated as:
\begin{equation}
    \mathcal{L}_{\textrm{\red{image}}}(\theta) = P_\theta(h_{vm}|h_{vv}, h_w),
\end{equation}
where ${h}_{vm}$ represents the masked visual tokens extracted by the ResNet backbone, ${h}_{vv}$ refers to the unmasked tokens, and $h_w$ corresponds to the word tokens of the entire report. 


\subsection{Visual-Linguistic Causal Intervention}
After the \red{RadCARE}, the learned visual-linguistic feature encoders still contain visual and linguistic biases from cross-modal confounders~\cite{yang2021deconfounded}. 
Therefore, \red{we further employ visual-linguistic causal intervention (VLCI) to discover the causal effect between visual and linguistic modalities during RRG}, as shown in Fig.~\ref{fig:method_overview} (b).

\subsubsection{Preliminaries}
\begin{figure}[t]
  \centering
  \includegraphics[width=1\linewidth]{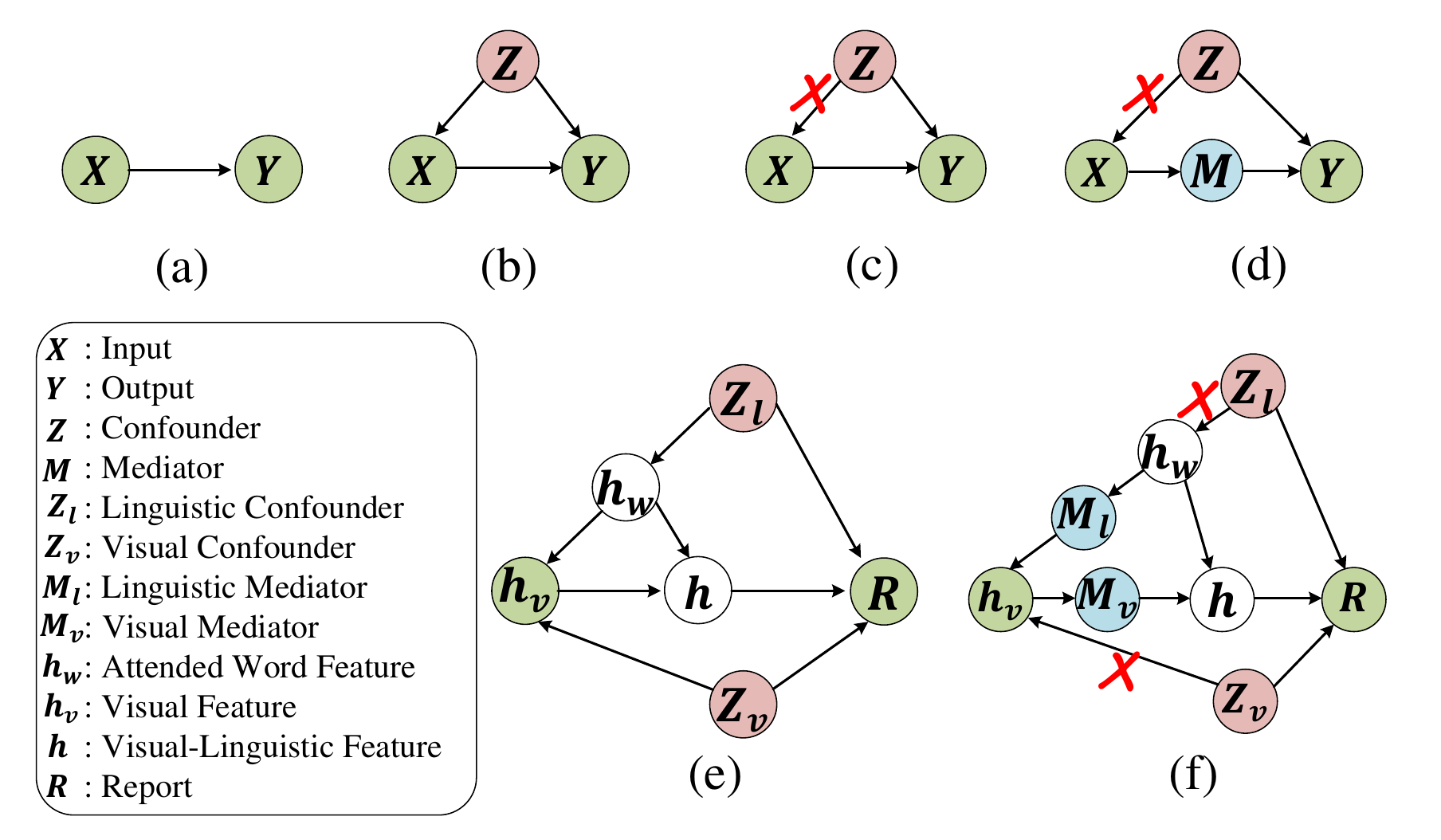}

  \caption{
  {The structural causal model (SCM) in (a) illustrates that $Y$ is directly caused by $X$ and indirectly influenced by confounders $Z$, as shown in (b). To address the confounding effect of $Z$, either by estimating the observable confounders $Z$ or by introducing a mediator $M$, the back-door and front-door interventions can be applied to block the path from $Z \to X$, as demonstrated in (c) and (d). In our proposed approach (e), we decompose the cross-modal confounders into visual ($Z_v$) and linguistic ($Z_l$) components. The front-door causal intervention is implemented by the mediators $M_v$ and $M_l$, effectively blocking the paths $Z_v \to h_v$ and $Z_l \to h_w$ in (f).}}

  \label{fig:method_scm}
\end{figure}

To clarify the mechanism of causal intervention, we introduce Pearl’s Structural Causal Model (SCM)~\cite{pearl2016causal}, as shown in Fig.~\ref{fig:method_scm}. The SCM is a mathematical framework used in causal inference to model the relations among variables and determine cause and effect relations. The SCM uses directed acyclic graphs (DAGs) to represent causal systems, where each variable is a node and the arrows represent causal relations between the variables.

In Fig.~\ref{fig:method_scm}(a), the chain structure $X \to Y$ represents the output $Y$ is affected by the input $X$, formulated as $P(Y|X)$. But the confounders $Z$ caused by data bias would lead to a spurious correlation, as shown in Fig.~\ref{fig:method_scm}(b). 
This graph indicates that two causes (both X and Z) lead to the common effect or outcome (Y), such as the cross-modal confounders of ``Normal Heart" and cross-modal feature of ``Pleural" leading to a confounded estimation of ``No Effusion" (the spurious correlation is ``Normal Heart" and ``No Effusion"). In this case, we formulate $P(Y|X)$ as:
\begin{equation}
\begin{aligned}
    P(Y|X) = \sum_{z} P(Y|X, Z=z)P(Z=z|X),
\end{aligned}
\end{equation}
where the confounders $Z$ generally brings about the observational bias via $P(z|X)$~\cite{liu2022show}.
To alleviate this issue, back-door causal intervention can be implemented by introducing the do calculus $do(\cdot)$~\cite{liu2022show, liu2023cross, lopez2017discovering, qi2020two}. \red{In Fig.~\ref{fig:method_scm} (c), w}e can calculate the probabilities of observable confounders and block the back-door path $Z \to F$, the interventional probability is as follows:
\begin{equation}
\begin{aligned}
    P(Y|do(X))=\sum_{z} P(Y|X, Z=z)P(Z=z),
    \label{eq:back_door_do}
\end{aligned}
\end{equation}
where $Z$ can be the learned RoI features of the heart and the attended word feature of ``Normal Heart". \red{Based on the conditional probabilities following intervention, it can estimate the true causal effects of each factor in the absence of confounding influences. These individual causal effects can then be aggregated to derive the overall causal effect under the given input conditions.} However, the back-door causal intervention is limited by the observability of confounders, and the front-door causal intervention provides a method of deconfounding in Fig.~\ref{fig:method_scm} (d). To eliminate the unobservable confounder, we introduce the mediator $M$ to cut off the link $X \gets Z \to Y$. The total probability $P(Y|do(X))$ can be represented as the following summation:
\begin{equation}
\begin{aligned}
    P(Y|do(X){)}=\sum_{m}P(Y|do(X),M=m)P(M=m|do(X)),
    \label{eq:front_door_do_1}
\end{aligned}
\end{equation}
where $M$ is introduced by $X$ without the back-door path. Thus, the intervention probability is equal to the conditional probability in the path $X \to M$~\cite{liu2023cross}. Besides, there is no direct causal path between $X$ and $Y$. In this way, the introduced summation in Eq.~(\ref{eq:front_door_do_1}) can be reformulated as:
\begin{equation}
\begin{aligned}
    &P(Y|do(X){)}\\
    &=\sum_{m}P(Y|do(X), do(M=m))P(M=m|X=x)\\
    &=\sum_{m}P(Y|do(M=m))P(M=m|X=x).
    \label{eq:front_door_do_2}
\end{aligned}
\end{equation}
To estimate $P(Y|do(M=m))$, we can apply the back-door intervention to cut off the link $M \gets X \gets Z \to Y$~\cite{liu2022show}. Therefore, we have the intervention probability formulation:
\begin{equation}
\begin{aligned}
    &P(Y|do(M=m))\\
    &=\sum_{\hat{x}}P(Y|do(M=m),X=\hat{x})P(X=\hat{x}|do(M=m))\\
    &=\sum_{\hat{x}}P(X=\hat{x})P(Y|X=\hat{x},M=m),
    \label{eq:front_door_do_3}
\end{aligned}
\end{equation}
where $\hat{x}$ is the selected features from $X$ not caused by $M$. Finally, we can calculate Eq.~(\ref{eq:front_door_do_2}) by applying Eq.~({\ref{eq:front_door_do_3}}):
\begin{equation}
\begin{aligned}
    &P(Y|do(M=m))=\\
    &\sum_{m}P(M=m|X=x)\sum_{\hat{x}}P(X=\hat{x})P(Y|X=\hat{x},M=m).
    \label{eq:front_door_do}
\end{aligned}
\end{equation}
However, existing front-door causal intervention methods employ diverse approaches for estimating mediators and conducting interventions. 
In {RRG}, we construct the structural causal model (SCM) similar to Image Caption~\cite{liu2022show}, assuming that $h_v$ is the visual feature, $h_w$ is the linguistic feature of the attended word, and $h$ is the fusion feature from $h_v$ and $h_w$ ($h_v \to h \gets h_w$), leading to the generation of report $R$. 
Moreover, $h_v$ would be influenced by $h_w$ through cross-attention and makes $h_v \gets h_w$, as shown in Fig.~\ref{fig:method_scm} (e).

\red{Previous method  ~\cite{liu2022show} employed a pre-constructed semantic dictionary for backdoor intervention, without accounting for complex representations found in radiological imaging. Differently, we construct mediators internally within the model to simulate front-door causal interventions, as illustrated in Figure~\ref{fig:method_vlci}. We further mine and construct mediators separately from visual and linguistic aspects by leveraging semantic correlations obtained during the pre-training phase. Finally, we estimate the actual causal effects by conditioning each mediator through mediator-based sub-distributions. In contrast, the baseline merely relies on the confounded distribution for estimation. To address this, we introduce the Front-Door Intervention Module (FIM) based on Equation (\ref{eq:front_door_do}) to disentangle the effects and construct the Visual Mediator Module (VDM) and Linguistic Mediator Module (LDM) separately.}



\subsubsection{Front-door Intervention Module (FIM)}

\begin{figure}[t]
 \centering
    \includegraphics[width=1\linewidth]{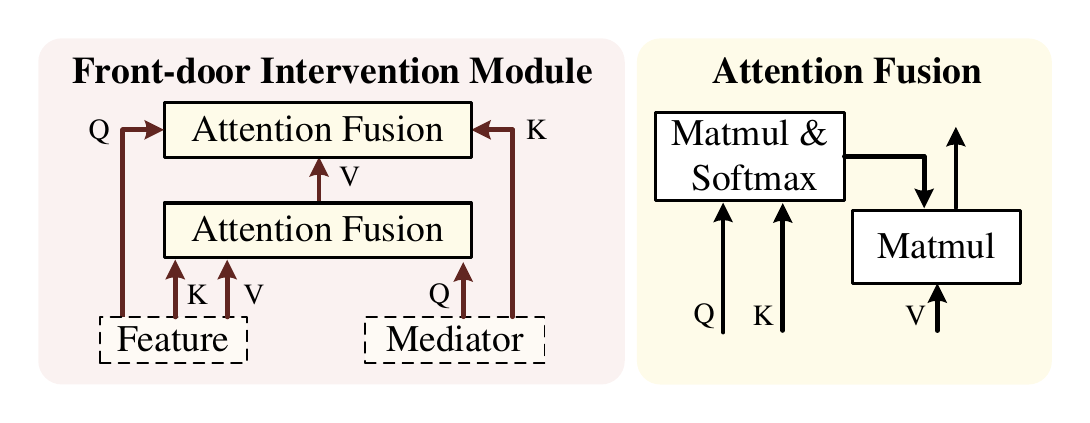}
 
    \caption{Illustration of the Front-door Intervention Module (FIM), which consists of two Attention Fusion layers with non-parameters.
    }
    \label{fig:method_fim}
\end{figure}

In Fig.~\ref{fig:method_scm} (e), the causal effects $h_v\to R$ and $h_w\to R$ are affected by the confounders $Z=\{Z_v,Z_l\}$ from back-door paths $h_v \gets Z_v\to R$ and $h_w \gets Z_l\to R$  ~\cite{liu2022show}, respectively. In our SCM, the non-interventional prediction can be expressed as:
\begin{equation}
\begin{aligned}
    &P(R|I) = P(R|h_v, h_w)\\
    &=\sum_{i=1}^{n}\sum_{z} P(w_{i}|h_v, h_w, Z=z)P(Z=z|h_v, h_w),
    \label{eq:mrg_non_do}
\end{aligned}
\end{equation}
where $Z$ brings the spurious correlation via $P(Z=z|h_v, h_w)$, leading to incorrect reports.
$h_v$ is the visual token from visual embedding, and $h_w$ is the linguistic token of the attended word from linguistic embedding.

Taking Fig.~\ref{fig:intro_bias_info} as an example, when $P(Z=\textrm{``Normal~Heart''}|h_v=\textrm{``Heart"}, h_w=\textrm{``Normal''})$ is large while $P(Z=\textrm{``Cardiomegaly"}|h_v=\textrm{``Heart''}, h_w=\textrm{``Normal"})$ is small in the training set, it tends to enlarge $P(R=\textrm{``No~Effusion"}|h_v, h_w, Z=\textrm{``Normal~Heart"})$ in the testing set. 

To mitigate visual-linguistic confounders and uncover the true causal structure, we apply causal front-door intervention by introducing mediator $M_v$ and $M_l$, respectively, as shown in Fig.~\ref{fig:method_scm} (f).
Generally, $Z_v$ is unobservable without a well-trained object detector, and the back-door path $h_v\gets Z_v \to R$ can be blocked by $M_v$ via learning the true causal effect $h_v \to M_v \to h\to R$. Similarly, the intervention on the back-door path $h_v\gets h_w\gets Z_l\to R$ can be implemented by calculating the $M_l$ without well-constructed confounders dictionaries. Thus, we can formulate Eq.~(\ref{eq:mrg_non_do}) as:
\begin{equation}
\begin{aligned}
    P(R|do(I))=P(R|do(h_v), do(h_w))
    \label{eq:mrg_do}
\end{aligned}
\end{equation}

To further estimate Eq.~(\ref{eq:mrg_do}) with the deep learning framework using Eq.~(\ref{eq:front_door_do}), we adopt Normalized Weighted Geometric Mean (NWGM)~\cite{xu2015show}
as:
\begin{equation}
\begin{aligned}
    P(R|do(h_v), do(h_w)) \approx \textrm{Softmax}(g(h_w, h_v, \hat{M_v}, \hat{M_l})).
    \label{eq:done}
\end{aligned}
\end{equation}
where $g(\cdot)$ denote the network mapping functions {including the transformer decoder and a linear layer}, $\hat{M_v}$ and $\hat{M_l}$ denote the estimations of $M_v$ and $M_l$ via VDM and LDM. 
In Fig.~\ref{fig:method_fim}, the FIM consists of two Attention Fusion layers and it is integrated into VDM and LDM. Different from the cascaded transformer~\cite{nguyen2022grit}, FIM does not have optimized parameters. It relies on extracted features and intervenes using estimated mediators of the corresponding modalities. This approach ensures that the estimation of mediators is entirely dependent on the sampling in VDM and LDM rather than FIM, achieving learning of mediators.


\begin{figure}[t]
 \centering
    \includegraphics[width=1\linewidth]{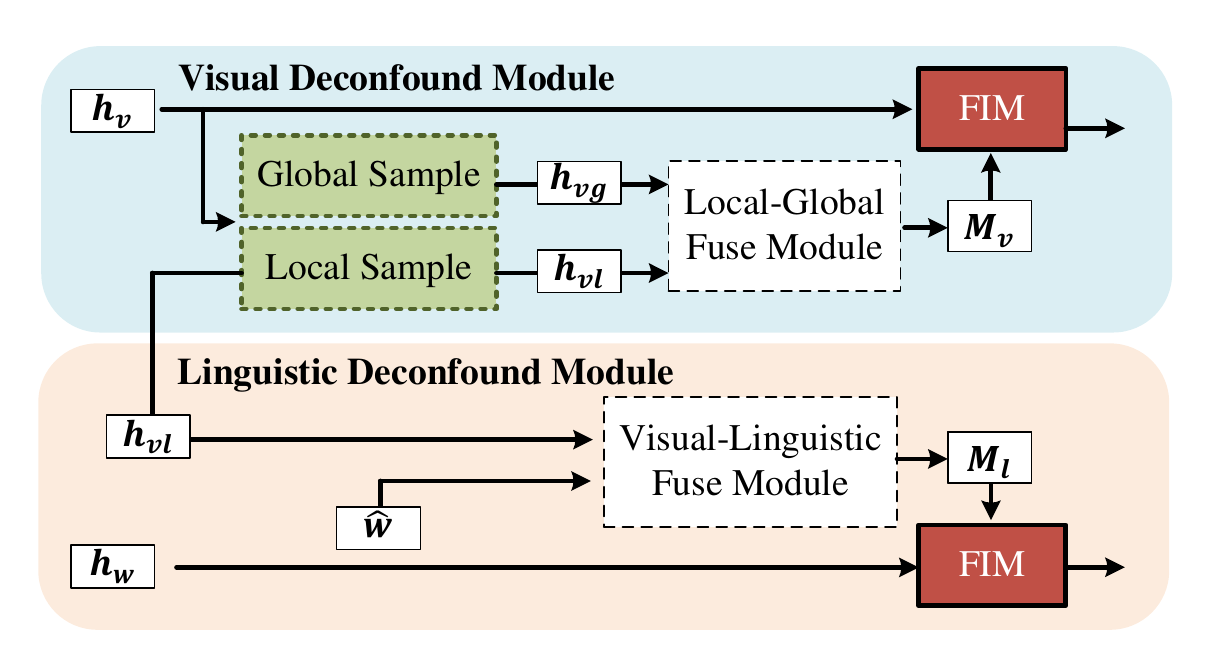}

    \caption{Illustration of the Visual Deconfound Module (VDM) and linguistic Deconfound Module (LDM).
    }
    \label{fig:method_vdm_ldm}
\end{figure}

\subsubsection{Visual Deconfounding Module (VDM)}
In Fig.~\ref{fig:method_vdm_ldm}, the visual mediator $M_v$ is calculated using local features $h_{vl}$ obtained from local sampling and global features $h_{vg}$ obtained from global sampling.
The features $h_{vl}$ represent local details acquired from Local Sampling, while $h_{vg}$ represents contour and position features obtained from Global Sampling~\cite{sun2022lesion}.
For example, the contour of the heart influences the determination of pleural effusion, and the texture of the lungs can also serve as a basis for detection.

\noindent\textbf{Local Sampling.} 
\red{We leverage the attention accumulated from the encoder to select the top $k$ tokens~\cite{he2022transfg}.} These selected visual tokens with high attention correspond to the report's keywords as $h_{vl}\in\mathbb{R}^{k\times d}$, where $k=6$ for each attention head, and $d$ is the dimension of the transformer. Subsequently, $h_{vl}$ is further enhanced using CaaM \cite{wang2021causal}, which excavates the local internal relations. Specifically, these highly attended tokens are computed not only with self-attention but also with negative attention scores, aiming to achieve more robust features. The purpose of $h_{vl}$ is to capture crucial local details in the image, which serve as the key basis for further processing.


\noindent\textbf{Global Sampling.} The global sampling is implemented by Down Sampling Transformer block, in which the $14\times 14$ visual tokens are down-sampled to $7\times 7$ as $h_{vg}\in\mathbb{R}^{49\times d}$. Max pooling in this block can better retain the global structure information in the image as the general features of the data itself. We formulate the operation as follows:
\begin{equation}
    h_{vg} = W[\textrm{P}(h_v) + \textrm{Attn}(\textrm{P}(\textrm{LN}(h_v))],
    \label{eq:dst}
\end{equation}
where P is the 2d max pooling layer, LN is layer normalization, Attn is the 2d relative attention \cite{dai2021coatnet}, and $W$ denotes the weights of the linear layer.

\noindent\textbf{Local-Global Fuse Module.} 
Finally, the $h_{vl}$ is integrated with $h_{vg}$ to enhance local details with global structural information via Local-Global Fuse Module formulated as Eq.~(\ref{eq:LGFM}), namely mediator $M_v$. 
\begin{equation}
    M_v=\textrm{FFN}([\textrm{MHA}(h_{vl}, h_{vl}, h_{vl}), \textrm{MHA}(h_{vl}, h_{vg}, h_{vg})])
    \label{eq:LGFM}
\end{equation}
where MHA and FFN are the Multi-Head Attention layer and Feed-Forward Network layer, respectively. $[\cdot,\cdot]$ denotes concatenation.

\begin{figure}[t]
  \centering
  \includegraphics[width=1\linewidth]{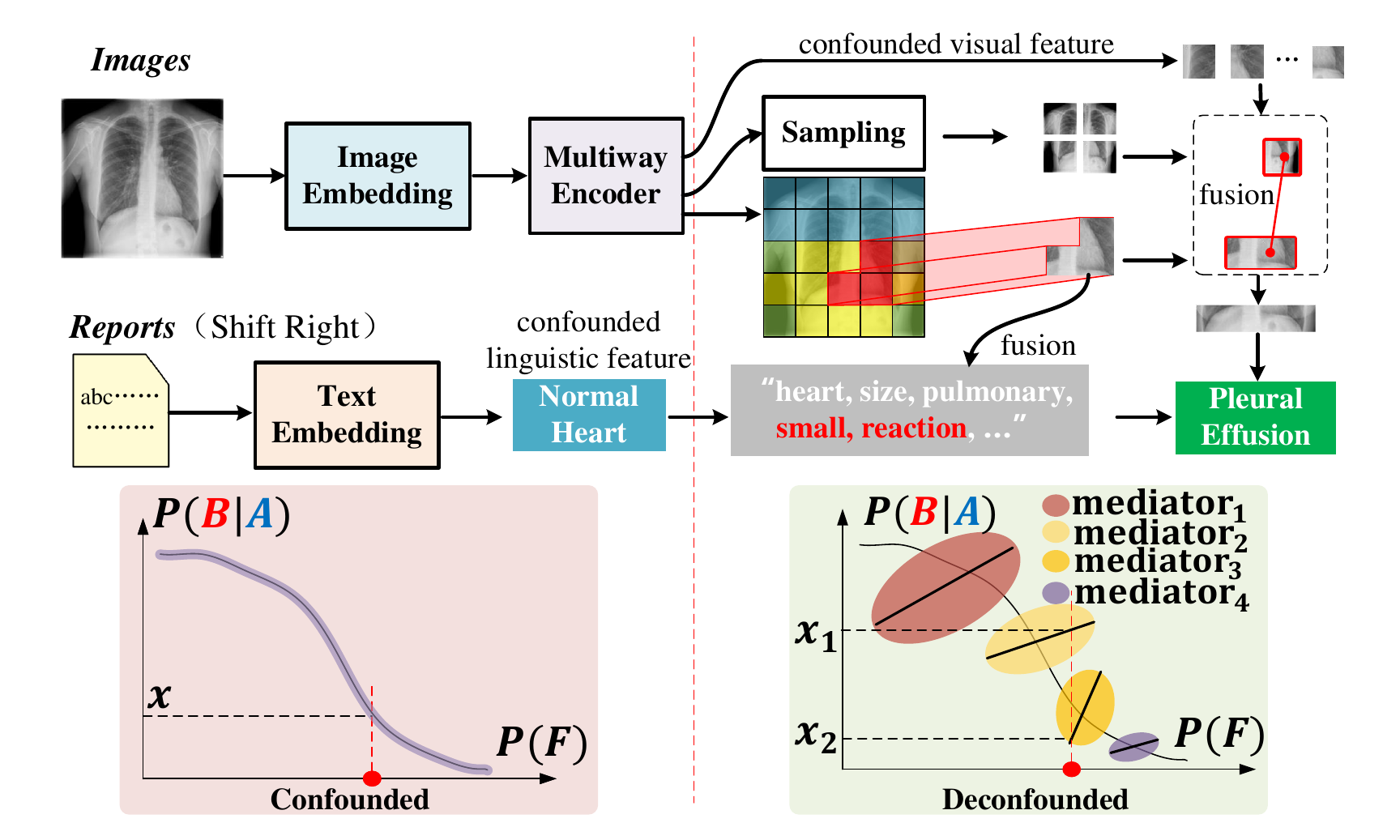}

  \caption{
  \red{Demonstration of the causal intervention mechanism, we integrate visual local-global features as visual mediators and combined visual local features with the vocabulary as linguistic mediators. Through their sub-distributions, we further estimate the causal effects in the RRG task.}}
 
  \label{fig:method_vlci}
\end{figure}

\subsubsection{Linguistic Deconfounding Module (LDM)}
For linguistic deconfounding, we have some observations from $h_v \gets h_w\gets Z_l\to R$ (Fig.~\ref{fig:method_scm} (e)): (1) the linguistic contexts can affect the generation of the next word, and (2) the attended word features affect the attended visual features via cross-attention~\cite{liu2022show}.
Additionally, the difference in word frequency brings a large distance deviation in embedding space, so the distance of word vectors cannot represent semantic relevance well~\cite{li2020sentence}. 

\noindent\textbf{Visual-Linguistic Fuse Module.} 
Thus, we calculate the linguistic mediator $M_l$ in embedding space via all vocabularies from the tokenizer as the global features and use $h_{vl}$ obtained from the VDM, which estimates the current word frequency to adjust the distribution of $h_w$, see in Fig.~\ref{fig:method_vdm_ldm}. 
\begin{equation}
\begin{aligned}
   &h^{'}_{vl} = \textrm{FFN}(\textrm{MHA}(h_{vl}, \hat{w}, \hat{w})); \\
   &M_t = \textrm{FFN}(\textrm{MHA}(h^{'}_{vl}, h_{vl}, h_{vl}))
\end{aligned}
\end{equation}
where $\hat{w}$ denotes all word tokens from the tokenizer. Then, we build the causal effect $ h_w \to M_l \to h_v \to M_v \to h \to R$ to cut off the back-door path $h_v \gets h_w\gets Z_l \to R$ via $M_l$.

In Fig.~\ref{fig:method_vlci}, the deconfounded visual and linguistic features are fed to the decoder to learn fused cross-modal features. The output layer is a linear projection with softmax operation, mapping probability into the dimensional equal to the vocabulary size. Finally, the training target is to minimize the negative log-likelihood loss according to Eq.~(\ref{eq:mrg_do}) and Eq.~(\ref{eq:done}):
\begin{equation}
\begin{aligned}
    &\mathcal{L}_{\textrm{nll}}(\theta) = -\sum_{i=1}^{n}log (\textrm{Softmax}(g(h_{w_{<i}}, h_v, \hat{M_v}, \hat{M_l})))
\end{aligned}
\end{equation}
where $n$ is the length of the report and $h_{w_{<i}}$ is the prefix text when estimating the word $w_i$.
\section{Experiment} 

\subsection{Experimental settings}

\begin{table}[]\renewcommand\arraystretch{0.9}
  \centering
  \setlength{\tabcolsep}{0.7mm}{
    \caption{The details of \red{CMCRL} and several comparison RRG models, the \#Enc and \#Dec denote the number of transformer layers in the encoder and decoder, respectively. 
    The marker $\clubsuit$ means 2 Contrastive Attentions, and $\spadesuit$ means Hierarchical LSTM. 
    The backbone of \red{CMCRL} is the first three blocks of Resnet101. 
    Besides, we show the employed model boosting modules, including the knowledge-aware module $\mathcal{K}$, template retrieval module $\mathcal{T}$, and memory-drive module $\mathcal{M}$.
    Additionally, we adopt two scales of the \red{CMCRL}, $\textbf{\red{CMCRL}}_3$ for the IU-Xray dataset and $\textbf{\red{CMCRL}}_6$ for the MIMIC-CXR dataset.
    }

  \label{tab:model_scale}
  \begin{tabular}{@{}lllccccc@{}}
    \toprule
    &Method              & Visual Embedding &  \#Enc  &   \#Dec & $\mathcal{K}$ & $\mathcal{T}$ & $\mathcal{M}$ \\
    \midrule
    \multirow{5}{*}{\rotatebox{90}{Light}}
    &R2Gen\cite{chen2020generating}               & Resnet101         & 3 & 3 &  &  & $\surd$\\
    &R2GenCMN\cite{chen2022cross}               & Resnet101         & 3 & 3 &  &  & $\surd$\\
    &CMCL\cite{liu2022competence}                & Resnet50          & - & $\spadesuit$ &  &  &  \\
    &PPKED\cite{liu2021exploring}               & Resnet152         & 2 & 1 & $\surd$ & $\surd$ &   \\
    &CA\cite{liu2021contrastive}                  & Resnet50          & $\clubsuit$ & $\spadesuit$ &   & $\surd$ &  \\
    &AlignTransformer\cite{you2021aligntransformer}    & Resnet50          & 3 & 3 & $\surd$  &   & $\surd$  \\
    &MMTN\cite{cao2023mmtn}                & DenseNet121              & 3 & 3 & $\surd$ &   & $\surd$ \\
    &\red{CAMANet~\cite{wang2024camanet}}&  \red{DenseNet121}              & \red{3} & \red{3} & &  &  \\
    &\red{SSVE~\cite{divya2024memory}}&  \red{ResNet101}              & \red{3} & \red{3} & &  & \red{$\surd$} \\
    \midrule
    \multirow{3}{*}{\rotatebox{90}{Heavy}}
    &M2TR\cite{nooralahzadeh2021progressive}                & Densenet151       & 6 & 12 &   &  & $\surd$\\
    &MGSK\cite{yang2022knowledge}             & Resnet101         & 12 & 3 & $\surd$ & $\surd$ &  \\
    &RAMT\cite{zhang2023semi}                & DenseNet121              & 6 & 6 & $\surd$ &   &  \\
    &DCL\cite{li2023dynamic}                & ViT Linear Projection           & 25 & 1 & $\surd$ & $\surd$  &  \\
    &\red{Med-LLM\cite{liu2024context}}& \red{MedCLIP-ViT} & \red{Q-former} & \red{32} & & \\
    \midrule
    &$\textbf{\red{CMCRL}}_3$                & Resnet101*        & 3 & 3 &  &  &  \\
    &$\textbf{\red{CMCRL}}_6$                & Resnet101*        & 6 & 6 &  &  &  \\
    \bottomrule
  \end{tabular}}

\end{table}

\subsubsection{Datasets}
\noindent \textbf{IU-Xray}~\cite{jamiaocv080}, also known as the Indiana University Chest X-ray Collection, is a publicly available radiological dataset widely used to evaluate the performance of {RRG} methods. It comprises 7,470 chest images and 3,955 corresponding reports, with an average report length of 30. To maintain consistency, we follow the setting used in R2Gen~\cite{chen2020generating} and partition the dataset into training, validation, and testing sets at a ratio of 7:1:2, ensuring that there is no overlap in patients. We tokenize the words with more than 3 occurrences and set the max length as 60. Note that we adopt two images (frontal and lateral views) as input for each sample.

\noindent \textbf{MIMIC-CXR}~\cite{johnson2019mimic} is a large-scale chest radiological dataset, with 377,110 images and 227,835 corresponding reports, of which the average length of reports is 48 in the training/val set and 61 in testing set. We use the official paradigm, the dataset is divided into the training set with 368,960 images and 222,758 reports, the validation set with 2,991 images and 1,808 reports, and the testing set with 5,159 images and 3,269 reports. Different from the IU-Xray dataset, MIMIC-CXR has instances of a single modality (only image or report) as well as multiple images corresponding to one report. Therefore, we use a single image as the input and tokenize the words with more than 10 occurrences, and set the max length as 80. 

\begin{table}[t]
  \centering
  \setlength{\tabcolsep}{1mm}{
    \caption{Comparison of computational cost between \red{CMCRL} with R2Gen and R2GenCMN. The $1^{st}$ and $2^{nd}$ best results are bolded and underlined, respectively.}

  \label{tab:model_cost}
  \begin{tabular}{@{}lllll@{}}
    \toprule
    Method          & Inference Time (s)  & Params (M) & FLOPs (TFLOPs) & BLEU-4\\
    \midrule
    R2Gen         & 97.14 & 78.07 & \textbf{3.66} & {10.3}\\ 
    R2GenCMN  & \textbf{40.25} & \textbf{58.65}& 12.98 & {\underline{10.5}}\\ 
    \red{CMCRL}       & \underline{97.10} & \underline{69.41} & \underline{10.50} & {\textbf{11.9}}\\ 
    \bottomrule
  \end{tabular}}
\end{table}

\begin{table*}
  \centering
  \setlength{\tabcolsep}{4mm}{
    \caption{The performances of \red{CMCRL} and other methods on IU-Xray and MIMIC-CXR datasets. The ${1}^{st}$ and ${2}^{nd}$ best results are bolded and underlined, respectively.
    For some methods, the results are missing and denoted by a ``-". 
  }

  \label{tab:main_result}
  \begin{tabular}{@{}lccccccccc@{}}
    \toprule
    Method             & BLEU-1 & BLEU-2 & BLEU-3 & BLEU-4  & Rouge-L & METEOR & Precision & Recall & F1\\
    \midrule
    \multicolumn{10}{c}{\textbf{IU-Xray Dataset}} \\
    \midrule
                    R2Gen\cite{chen2020generating}              & {47.0} & {30.4} & {21.9} & {16.5}    & {37.1} & {18.7} & - & - & - \\ 
                    CMCL\cite{liu2022competence}               & {47.3} & {30.5} & {21.7} & {16.2}    & {37.8} & {18.6} & - & - & - \\ 
                    PPKED\cite{liu2021exploring}              & {48.3} & {31.5} & {22.4} & {16.8}   & {37.6} & {19.0} & - & - & - \\
                    CA\cite{liu2021contrastive}                & {49.2} & {31.4} & {22.2} & {16.9}   & {38.1} & {19.3} & - & - & - \\
                    AlignTransformer\cite{you2021aligntransformer}   & {48.4} & {31.3} & {22.5} & {17.3}    & {37.9} & {\underline{20.4}} & - & - & - \\
                    M2TR\cite{nooralahzadeh2021progressive}               & {48.6} & {31.7} & {23.2} & {17.3}   & {\underline{39.0}} & {19.2} & - & - & - \\
                    MGSK\cite{yang2022knowledge}             & {\underline{49.6}} & {\underline{32.7}} & {\underline{23.8}} & {\underline{17.8}} & {38.1} & - & - & - & - \\

                    RAMT\cite{zhang2023semi}          & {48.2} & {31.0} & {22.1} & {16.5}  & {37.7} & {19.5} & - & - & - \\
                    MMTN\cite{cao2023mmtn}            & {48.6} & {32.1} & {23.2} & {17.5}  & {37.5} & - & - & - & - \\
                    DCL\cite{li2023dynamic} & - & - & - & {16.3} & {38.3} & {19.3} & - & - & - \\
                    \red{SSVE\cite{divya2024memory}} & \red{49.2} & \red{32.1} & \red{23.3} & \red{18.0} & \red{37.9} & \red{20.3} & \red{-} & \red{-} & \red{-} \\
                    \red{Med-LLM\cite{liu2024context}} & \red{-} & \red{-} & \red{-} & \red{16.8} & \red{38.1} & \red{20.9} & \red{-} & \red{-} & \red{-} \\
                    $\textbf{\red{CMCRL}}_3 \textbf{(ours)}$               & {\textbf{50.5}} & {\textbf{33.4}} &  {\textbf{24.5}} & {\textbf{19.0}} & {\textbf{39.4}} & {\textbf{21.0}} & - & - & - \\\hline
    \midrule
    \multicolumn{10}{c}{\textbf{MIMIC-CXR Dataset}} \\
    \midrule
                    R2Gen\cite{chen2020generating}                   & {35.3} & {21.8} & {14.5} & {10.3} &  {27.7}  & {14.2} & {33.3} & {27.3} & {27.6}\\ 
                    R2GenCMN\cite{chen2022cross}                     & {35.6} & {21.9} & {14.7} & {10.5} &  {27.8}  & {14.1} & {33.4} & {27.5} & {27.8}\\ 
                    CMCL\cite{liu2022competence}                     & {33.4} & {21.7} & {14.0} & {09.7} &  {28.1} & {13.3} & - & - & - \\ 
                    PPKED\cite{liu2021exploring}                     & {36.0} & {22.4} & {14.9} & {10.6} &  {\textbf{28.4}} & {14.9} & - & - & - \\
                    CA\cite{liu2021contrastive}                      & {35.0} & {21.9} & {15.2} & {10.9} &  {\underline{28.3}} & {15.1} & {35.2} & {29.8} & {30.3}\\
                    AlignTransformer\cite{you2021aligntransformer}   & {37.8} & {23.5} & {15.6} & {11.2} &  {\underline{28.3}} & {\underline{15.8}} & - & - & - \\
                    M2TR\cite{nooralahzadeh2021progressive}          & {37.8} & {23.2} & {15.4} & {10.7} &  {27.2} & {14.5} & {24.0} & {\textbf{42.8}} & {30.8}\\
                    MGSK\cite{yang2022knowledge}                     & {36.3} & {22.8} & {15.6} & {11.5} &  {\textbf{28.4}} & -           & {45.8} & {34.8} & {37.1} \\
                    RAMT\cite{zhang2023semi}                         & {36.2} & {22.9} & {15.7} & {11.3} &  {\textbf{28.4}} & {15.3}  & {38.0} & {34.2} & {33.5} \\
                    MMTN\cite{cao2023mmtn}                           & {\underline{37.9}} & {\underline{23.8}} & {\underline{15.9}} & {\underline{11.6}} & {\underline{28.3}} & {\textbf{16.1}} & - & - & - \\
                    DCL\cite{li2023dynamic}  & - & - & - & {10.9} & {\textbf{28.4}} & {15.0} & {\underline{47.1}} & {\underline{35.2}} & {\underline{37.3}}\\
                    \red{CAMANet\cite{wang2024camanet}} & \red{37.4} & \red{23.0} & \red{15.5} & \red{11.2} & \red{27.9} & \red{14.5} & \red{48.3} & \red{32.3} & \red{38.7} \\
                    $\textbf{\red{CMCRL}}_6 \textbf{(ours)}$              & {\textbf{40.0}} & {\textbf{24.5}} & {\textbf{16.5}} & {\textbf{11.9}} & {28.0} & {15.0} & {\textbf{48.9}} & {34.0} & {\textbf{40.1}}\\
    \bottomrule
  \end{tabular}}
\end{table*}

\subsubsection{Baseline Models}
We compare the proposed \red{CMCRL} model with several state-of-the-art {RRG} models, including R2Gen \cite{chen2020generating}, R2GenCMN \cite{chen2022cross}, CMCL \cite{liu2022competence}, PPKED \cite{liu2021exploring}, CA \cite{liu2021contrastive}, AlignTransformer \cite{you2021aligntransformer}, M2TR \cite{nooralahzadeh2021progressive}, RAMT \cite{zhang2023semi}, MMTN \cite{cao2023mmtn}, MGSK \cite{yang2022knowledge}, DCL \cite{li2023dynamic}, \red{SSVE\cite{divya2024memory}, CAMANet\cite{wang2024camanet}, and Med-LMM\cite{liu2024context}}. These models are categorized into lightweight and heavyweight models. The lightweight models comprise no more than 3-layer modules for both the encoder and the decoder, as detailed in Table~\ref{tab:model_scale}. Most of these models incorporate various modules to enhance their performance, such as the knowledge-aware module, template retrieval module, and memory-drive module, which can be computationally expensive.  It is important to note that the total parameters of our \red{CMCRL} are less than the R2Gen, while \red{CMCRL} is more efficient by eliminating the recursive memory calculation dependency, as shown in Table~\ref{tab:model_cost}.

\subsubsection{Evaluation Metrics} {We adopt the widely used natural language generation (NLG) metrics, including 
BLEU~\cite{papineni2002bleu}, a metric evaluates how similar the generated text is to reference texts by calculating n-gram precision.
ROUGE-L~\cite{rouge2004package} evaluates both precision and recall and also considers synonyms, stemming, and paraphrasing.
METEOR~\cite{banerjee2005meteor} focuses on the longest common subsequence between the generated text and the reference text, capturing fluency and coherence.
CIDEr~\cite{vedantam2015cider} considers term frequency-inverse document frequency (TF-IDF) weighting to reduce the impact of overly common words. 
Since the {RRG} specifically focuses on the abnormality detection accuracy rather than the text fluency and similarity with the real report, we further adopt clinical efficacy (CE) metrics~\cite{chen2020generating, liu2021contrastive, nooralahzadeh2021progressive, yang2022knowledge}. It is calculated by the labels extracted from CheXpert~\cite{irvin2019chexpert}. 
Specifically, the extracted positive labels are considered as positives, while the non-positive labels (negative, not mentioned, and uncertain) are treated as negatives. Using this approach, we calculate the micro-precision, micro-recall, and micro-F1 scores between the labels from reference reports and generated reports.}

\subsubsection{Implementation Settings} We use the first three blocks of ResNet101~\cite{he2016deep} to extract 1,024 feature maps, which are projected into 512 maps of size $14 \times 14$. 
The dimension of the transformer layers and the number of attention heads are fixed to 512 and 8, respectively. 
In the IU-Xray dataset (referred to as \red{CMCRL}$_3$), both the encoder and decoder consist of 3 layers, whereas in the MIMIC-CXR dataset (referred to as \red{CMCRL}$_6$), both the encoder and decoder comprise 6 layers, as illustrated in Table~\ref{tab:model_scale}. The variation in the number of layers can be attributed to the significantly larger size of the MIMIC-CXR dataset in comparison to the IU-Xray dataset.
During pre-training, we utilize a dataset that combines IU-Xray and MIMIC-CXR datasets, resulting in 4,347 unique words. We perform fine-tuning on these two datasets separately with the same tokenizer as \red{RadCARE}. The batch size is set to 64 during pre-training and 16 during fine-tuning.
In the pre-training stage, we adopt an image mask rate of 85\% for \red{masked image restoration}.  
The \red{RadCARE} is trained using the AdamW optimizer with a warm-up step of 10\% of the total training steps, and the peak learning rate is set to 5e-4.  The weight decay of the optimizer is set to 1e-2, and the total epochs are set to 30 in pre-training.
In the fine-tuning stage, the model is fine-tuned using the Adam optimizer with an initial learning rate of 1e-5 and a weight decay of 5e-5 for 10 epochs on both the IU-Xray and MIMIC-CXR datasets.

\subsection{Quantitative Analysis}
\subsubsection{NLG Metric}
As shown in Table~\ref{tab:main_result}, our \red{CMCRL} outperforms nearly all the {RRG} methods. Specifically, compared with the lightweight model MMTN~\cite{cao2023mmtn}, \red{CMCRL}$_3$ significantly improves the BLEU-1 metric by 1.9\% on the IU-Xray dataset. 
Similarly, when compared with the heavyweight model RAMT~\cite{zhang2023semi}, \red{CMCRL}$_6$ achieves a substantial 3.8\% boost in the BLEU-1 metric on the MIMIC-CXR dataset. This significant improvement in the BLEU-1 metric highlights our method's ability to select words more precisely.  Notably, on the IU-Xray dataset, our tokenizer incorporates words from the MIMIC-CXR dataset, leading to even greater estimation challenges.
To assess text fluency more effectively, we also consider the precise match of four consecutive words using the BLEU-4 metric. Our \red{CMCRL} demonstrates a 1.5\% improvement over MMTN on the IU-Xray dataset and a 0.6\% improvement over RAMT on the MIMIC-CXR dataset.  Notably, the longer descriptions in the MIMIC-CXR dataset present visual-linguistic data bias and significant spurious correlations among multiple words. Our \red{CMCRL} effectively leverages cross-modal causal intervention to achieve performance improvement.

In contrast to BLEU, Rouge-L focuses more on the structural and sequential similarity of sentences. Our method demonstrates significant superiority on the IU-Xray dataset while exhibiting slightly lower performance compared to DCL~\cite{li2023dynamic} on the MIMIC-CXR dataset. However, Table~\ref{tab:main_result} reveals that all methods achieve Rouge-L scores around 0.280 on the MIMIC-CXR dataset. This observation suggests that, when generating long sentences, the differences in structure and sequence with the reference reports are similar across various methods. As a result, this metric may not be a sensitive evaluation criterion for this dataset. 

Similarly, the METEOR metric considers synonyms and phrase reordering, emphasizing diverse matches of phrases and vocabulary. As indicated in Table~\ref{tab:main_result}, \red{CMCRL} achieves the best performance in terms of the METEOR metric on the IU-Xray dataset, but it falls short compared to MMTN on the MIMIC-CXR dataset. This suggests that there are some limitations in matching phrase and vocabulary diversity in our approach. By referring to Table~\ref{tab:model_scale}, we observe that both the best-performing MMTN and AlignTransformer benefit from knowledge supplementation and the assistance of a memory module, potentially enhancing the model's ability to estimate synonyms more effectively. While our approach relies on learning semantics solely from the linear projection space of tokens, which may have limitations.


\subsubsection{CE Metric}
The purpose of {RRG} is to alleviate the burden on radiological professionals and provide precise diagnostic evidence. Therefore,  the CE metric is more appropriate for evaluating the clinical significance of {RRG} methods compared to NLG metrics, as it specifically assesses the accuracy of abnormality classification. 
The CE metric is only applied to the MIMIC-CXR dataset because the label extractor (CheXpert)~\cite{irvin2019chexpert} is specially designed for MIMIC-CXR to obtain class labels. Compared with the lightweight R2Gen in Table~\ref{tab:main_result}, \red{CMCRL} demonstrates a remarkable improvement of 15.6\% in Precision, 6.7\% in Recall, and 12.5\% in F1-Score. This validates that \red{CMCRL} can provide a more accurate clinic diagnosis rather than merely generating fluent reports. Additionally, our \red{CMCRL} outperforms the state-of-the-art method DCL in terms of Precision score and F1 score. Notably, when compared to several knowledge-based methods, our approach achieves a more efficient clinical evidence generation by relying solely on causal intervention without requiring additional knowledge assistance. Nevertheless, the Recall score is relatively lower compared to M2TR in Table~\ref{tab:main_result}, indicating that the dataset may contain extreme categories. M2TR employs staged generation, which leverages the preliminary concepts generated in the first stage, enabling effective anomaly detection. However, this process can also lead to an increase in false positives, consequently reducing the Precision score.

\begin{figure}[t]
  \centering
  \includegraphics[width=1\linewidth]{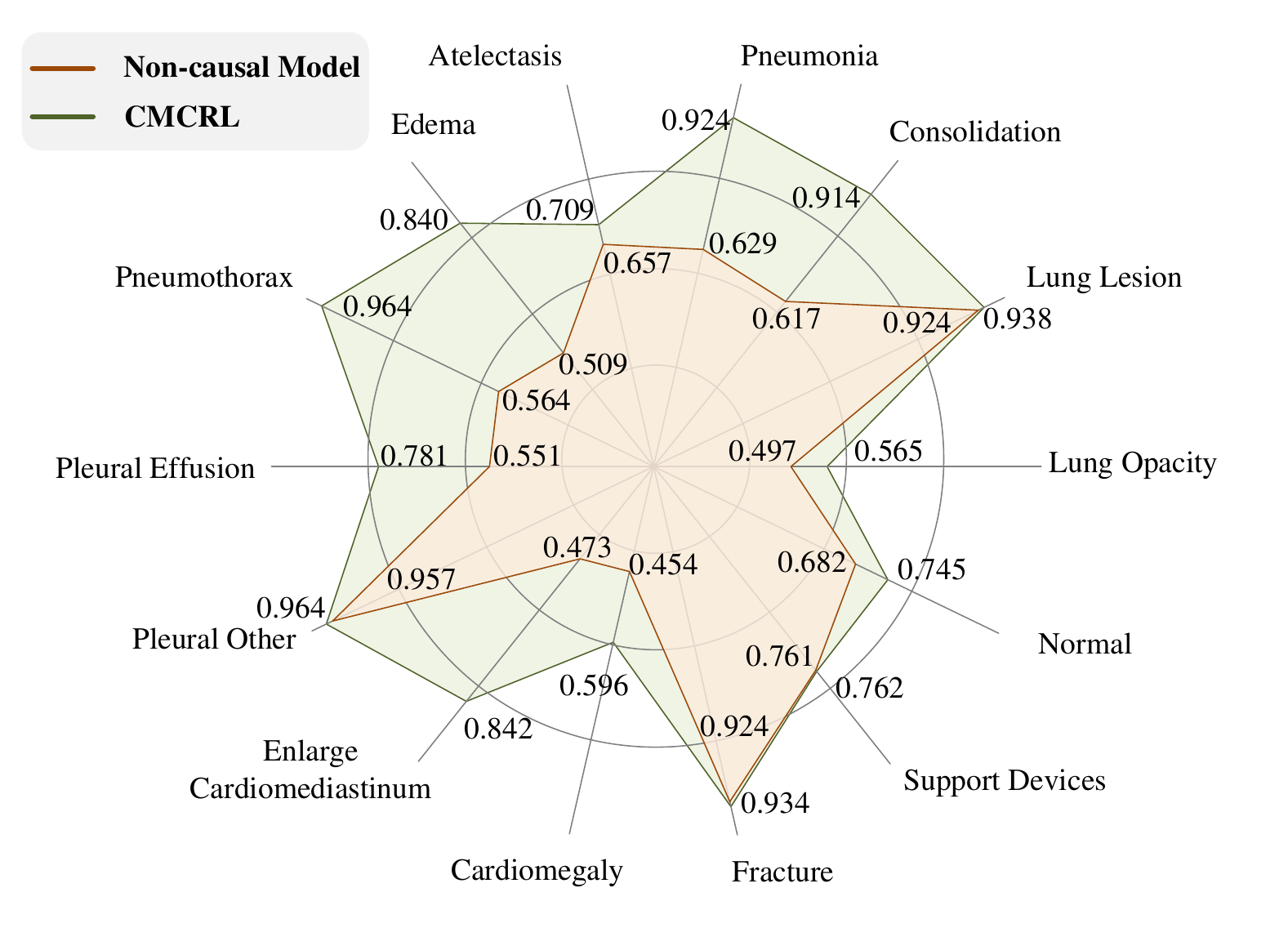}

  \caption{Evaluation of abnormality classification results (accuracy) on MIMIC-CXR. The baseline model is the transformer without causal intervention.}
  \label{fig:experiment_abnormality}
\end{figure}

\subsection{Qualitative Analysis}

\subsubsection{Abnormalities Detection}
To further validate the assistance of causal intervention (i.e., alleviate the
burden on radiological professionals and provide precise diagnostic evidence) in our method, we extract 14 categories of labels from the reports generated by the baseline and \red{CMCRL}, and evaluated their accuracy, as shown in Fig.~\ref{fig:experiment_abnormality}. Our approach achieves significant performance improvements in all categories, particularly in ``Edema" ($0.509 \to 0.840$) and ``Enlarge Cardiomediastinum" ($0.473 \to 0.842$). This is because our \red{CMCRL} explores sufficient visual information and further produces more accurate and less biased descriptions by cross-modal causal intervention than the Transformer baseline. 
However, the estimation of some categories remains ambiguous, e.g., ``Lung Opacity". It reveals that \red{CMCRL} can provide a comprehensive consideration of various radiologic signs to detect the abnormality but give less improvement for the single source abnormality. For example, whether ``Edema" is caused by the heart has different radiologic signs, while the increase in lung density can be considered as ``Lung Opacity". Thus, \red{CMCRL} can capture the abnormality with complex causes more effectively, where exists more spurious correlations. Besides, Fig.~\ref{fig:experiment_abnormality} shows the unavailability of causal intervention in independent abnormalities, e.g. ``Support Devices".

\subsubsection{Causal Consistency}
\begin{figure}[!t]
 \centering
    \includegraphics[width=1\linewidth]{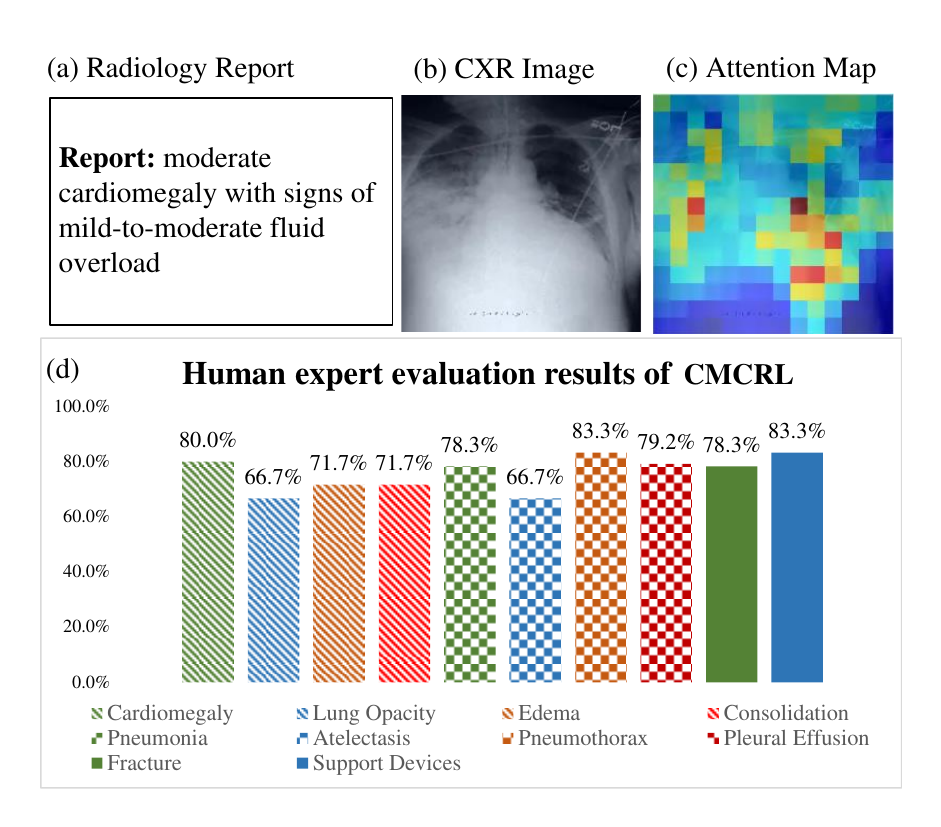}
    \caption{{
    A sample from the questionnaire. (a) is the generation report, (b) is the input image and (c) is the heatmap from cross-attention layer. (d) is the human expert evaluation results of \red{CMCRL}.
    }}
    \label{fig:stat_result}
\end{figure}

{To assess whether \red{CMCRL} utilized appropriate visual evidence for its inference, we develop a questionnaire and distributed it to human experts for evaluation. We select 10 data categories, each containing 10 samples, resulting in a total of 100 instances. Visual attention maps are generated based on the descriptions in the reports, where attention is captured through the cumulative response of entire sentences. These visualizations are compiled into a questionnaire (as shown in Figure \ref{fig:stat_result} (a-c)) and distributed to 12 experienced radiological professionals. They evaluate the plausibility of the results based on the original images, attention maps, and corresponding sentences.}

{As shown in Fig \ref{fig:stat_result} (d), our approach achieves an overall reasonableness rating of 76.3\%, indicating that most of the samples are accepted by experts. Moreover, due to the low image resolution (224*224), the experts find it challenging to make judgments. Furthermore, the results for Cardiomegaly, Pneumothorax, and Support Devices exhibited significant consistency with human priors. This underscores the effectiveness of our approach in uncovering latent causal relations rather than relying on spurious correlations for judgment. Additionally, we intend to further advance the field in interpretable and trustworthy reasoning, with the goal of leveraging explicit causal relations for inference in the future.}

\subsubsection{Case Study}
\begin{figure*}[h]
 \centering
    \includegraphics[width=1\linewidth]{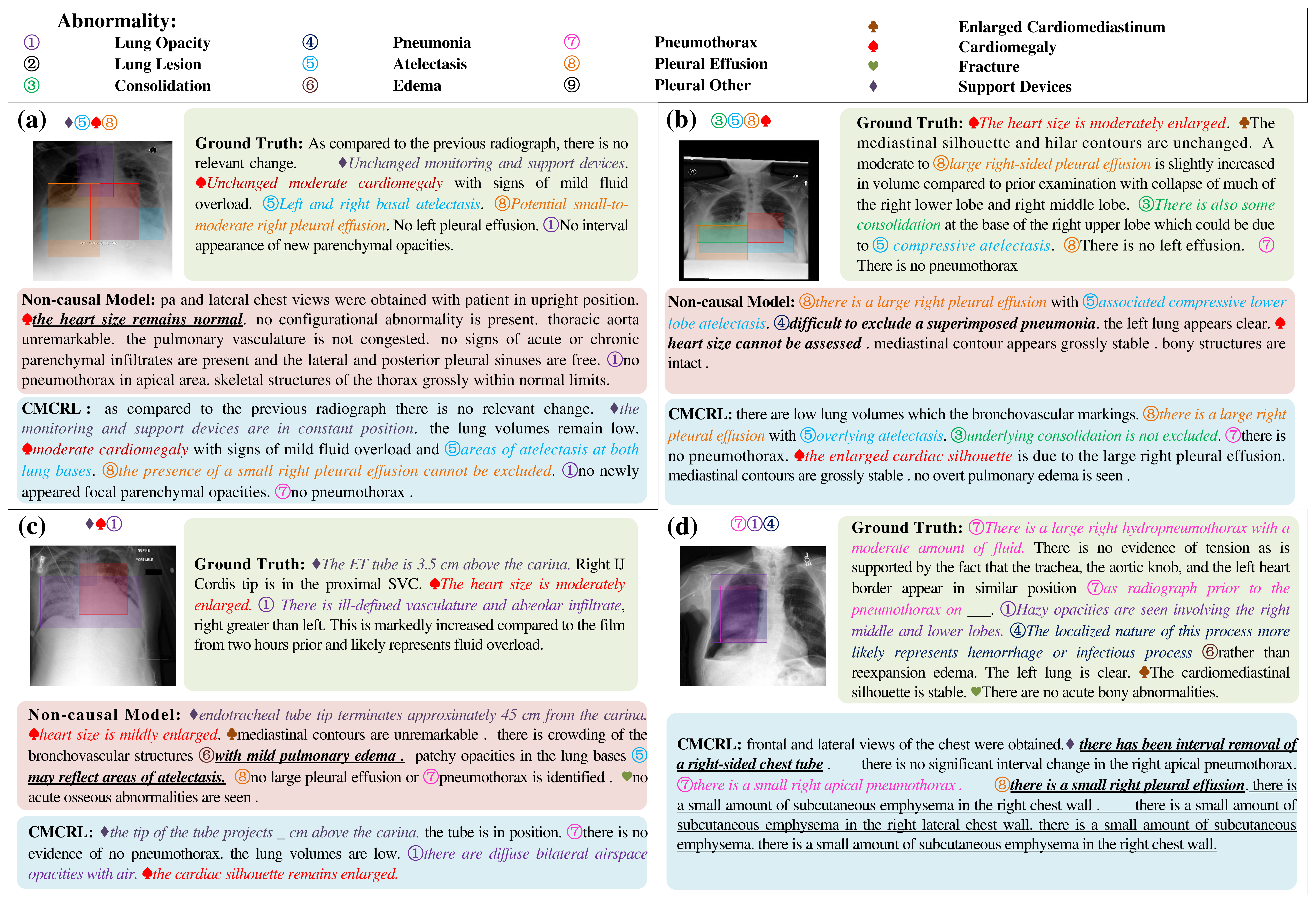}
    \caption{{The results of the Non-causal Model (Baseline) and \red{CMCRL} models on the MIMIC-CXR dataset are presented in (a-c), and (d) shows the false sample in \red{CMCRL}. Thirteen kinds of abnormalities are marked with different markers and colors. Note that keywords in the reports are also marked with different markers and colors. Correctly identified abnormalities are marked in the corresponding color, while other descriptions in bold, italics, and underscores are incorrect. Descriptions marked only with underscores indicate repeated words. 
    }}
    \label{fig:experiment_result}
\end{figure*}

We further conduct the qualitative analysis on the MIMIC-CXR dataset via three intuitive generated examples of the baseline and the \red{CMCRL} in Fig.~\ref{fig:experiment_result}.
In Fig.~\ref{fig:experiment_result} (a), the reference report contains four abnormalities. However, the baseline model neglects all abnormalities, while \red{CMCRL} accurately identifies all abnormalities. This indicates that our VDM comprehensively captures all essential visual features, crucial for {RRG}.
Fig.~\ref{fig:experiment_result} (b) shows an example where the same visual region is simultaneously discovered by the baseline and the \red{CMCRL} but leads to different descriptions. Our \red{CMCRL} can accurately describe the heart, while the baseline is uncertain and even has a miscalculation of pneumonia. 
It shows that LDM can alleviate the semantic bias caused by word frequency in word embedding space.
Fig.~\ref{fig:experiment_result} (c) illustrates a complex causal graph, where ``atelectasis" and ``edema" could also be causes of ``cardiomegaly." However, the baseline fails to correctly consider the causes of ``cardiomegaly" and erroneously captures these pieces of evidence. In contrast, \red{CMCRL} leverages the causal intervention module to disentangle these confounders, enabling careful consideration of various pieces of evidence for an accurate judgment.

In Fig.~\ref{fig:experiment_result} (d), the ground truth indicates the presence of hydropneumothorax, a condition characterized by the simultaneous presence of gas and fluid in the chest, whereas pleural effusion contains only fluid. Although \red{CMCRL} correctly identifies the presence of gas and fluid in the chest, leading to the diagnosis of pneumothorax, it erroneously estimates pleural effusion due to insufficient relevant knowledge. In this example, our method produces inaccurate text and fails to identify pneumonia and lung consolidation. Moreover, while VDM and LDM excel at recognizing visual and language concepts, detecting highly specialized concepts with latent relations not explicitly present in the data presents challenges.

\begin{table}
  \centering
  \setlength{\tabcolsep}{2mm}{
    \caption{Ablation Result of DenseNet121 backbone.}

  \label{tab:ablation_densenet}
  \begin{tabular}{@{}lllll@{}}
    \toprule
    Method on IU-Xray          & BLEU-4  & Rouge-L & METEOR\\
    \midrule
    ResNet101         & {14.8} & {34.5} & {18.0}\\ 
    ResNet101 +  VLCI  & {\textbf{19.0} (+4.2)} & {\textbf{39.4} (+4.9)}& {\textbf{21.0} (+3.0)}\\ 
    DenseNet121       & {16.4} & {36.1} & {18.3}\\ 
    DenseNet121 +  VLCI& {\textbf{17.6} (+1.2)} & {\textbf{39.4} (+3.3)} & {\textbf{19.5} (+1.2)}\\ 
    \midrule
    Method on MIMIC-CXR       & BLEU-4  & CIDEr & METEOR\\
    \midrule
    ResNet101& {10.1} & {13.0} & {13.5}\\ 
    ResNet101 +  VLCI& {\textbf{11.9} (+1.8)} & {\textbf{19.0} (+6)}& {\textbf{15.0} (+1.5)}\\ 
    DenseNet121& {9.1} & {11.1} & {12.7}\\ 
    DenseNet121 +  VLCI & {\textbf{11.3} (+2.2)} & {\textbf{13.4} (+2.3)} & {\textbf{14.9} (+2.2)}\\ 
    \bottomrule
  \end{tabular}}
\end{table}

\begin{table}[t]
  \centering
  \setlength{\tabcolsep}{3mm}{
  \caption{We evaluated the performance of various masking ratios for \red{masked image restoration} on the IU-Xray dataset. We pre-trained the \red{RadCARE} model for 100 epochs and then fine-tuned it in the baseline (non-causal model) for an additional 5 epochs.}

  \label{tab:experiment_vlp_mask}
  \begin{tabular}{@{}lcccc@{}}
    \toprule
    Masking Ratio           & BLEU-1  & BLEU-4  & CIDEr & Rouge-L \\
    \midrule
    Baseline            & {43.3} & {14.8} & {50.1} & {34.5} \\ 
    w/ 75\%             & {45.0} & {16.0} & {48.6} & {\textbf{36.0}} \\ 
    w/ 85\%             & {\textbf{45.2}} & {\textbf{16.1}} & {\textbf{52.2}} & {35.1} \\
    w/ 95\%             & {43.2} & {15.3} & {46.0} & {34.6} \\ 
    \bottomrule
  \end{tabular}}

\end{table}

\subsection{Ablation Studies}
\subsubsection{Effectiveness of VLCI}
In Table~\ref{tab:ablation_densenet}, we adopt the same setting as ResNet101 to conduct the ablation experiments using DenseNet121 as the backbone. We perform \red{RadCARE} and fine-tuning with causal intervention on IU-Xray and MIMIC-CXR datasets. Due to the small scale of the IU-Xray dataset and the simplicity of report content, the persuasiveness of the CIDEr metric on this dataset is limited. We evaluated BLEU-4, Rouge-L, and METEOR on the IU-Xray dataset, and the results indicate that our method still achieves a significant improvement with DenseNet121. However, we observed that the performance of models using DenseNet121 as the backbone is notably inferior to those using ResNet101. We speculate that while DenseNet121 is efficient in feature extraction due to its dense connections, it may not always provide the best performance for tasks involving specific types of radiological images or requiring deeper feature extraction capabilities. Additionally, we conducted experiments on the larger MIMIC-CXR dataset, using BLEU-4, CIDEr, and METEOR as evaluation metrics. The results further confirm our speculations and validate the effectiveness of our approach.

\begin{table}[!t]
  \centering
  \setlength{\tabcolsep}{1mm}{
    \caption{The performance of different pre-training methods on IU-Xray, the result marked by * means fine-tuning with 10 epochs, while the rest only use the encoder with 100 epochs. \red{``text" is denoted as postfix text generation and ``image" is masked image restoration}. The result marked by $\dagger$ is from \cite{Chen_2022_CVPR}.}
  \label{tab:ablation_vlp}
  \begin{tabular}{@{}lccc@{}}
    \toprule
    Method           & BLEU-4  & CIDEr & Rouge-L \\
    \midrule
    Baseline                  & {14.8} & {50.1} & {34.5} \\ 
    w/ MAE                    & {15.4} & {48.6} & {36.0} \\ 
    w/ VisualGPT$\dagger$     & {15.9} & {49.7} & {\textbf{37.4}} \\ 
    w/ MIM              & {16.2} & {\underline{60.2}} & {36.2} \\
    w/ \red{TEXT}              & {\underline{16.5}} & {53.8} & {\underline{36.5}} \\ 
    w/ \red{TEXT+IMAGE}          & {16.0} & {43.1} & {36.4} \\ 
    \midrule
    w/ \red{TEXT}*             & {15.1} & {39.9} & {34.9} \\ 
    w/ \red{TEXT+IMAGE}*         & {16.1} & {52.2} & {35.1} \\ 
    \textbf{w/ \red{RadCARE}}* (Ours)   & {\textbf{17.0}} & {\textbf{63.1}} & {36.3} \\ 
    \bottomrule
  \end{tabular}}
\end{table}

\begin{table}
  \centering
  \setlength{\tabcolsep}{0.7mm}{
    \caption{Ablation analysis of our \red{CMCRL}. The Baseline is implemented by the transformer. The marker at the Baseline (non-causal model) and R2Gen~\cite{chen2020generating} means the operation in the brackets.}
          \vspace{-10pt}
  \label{tab:ablation_baseline}
  \begin{tabular}{@{}llccc@{}}
    \toprule
    Dataset     & Method        & BLEU-4  & CIDEr & Rouge-L \\
    \midrule
                & Baseline                                             & {14.8} & {50.1} & {34.5} \\ 
                & Baseline$^{w\blacklozenge}$ (w/ VDM)                 & {16.0} & {52.1} & {36.4}\\
                & Baseline$^{w\bullet}$ (w/ LDM)                       & {15.5} & {50.9} & {36.1}\\
                & Baseline$^{w\blacklozenge \bullet} $ (w/ VDM\&LDM)   & {16.3} & {54.4} & {36.1} \\ 
                \cmidrule{2-5}
                & R2Gen                                                & {16.5} & {49.3} & {36.0} \\
                & R2Gen$^{w\blacklozenge}$ (w/ VDM)                    & {17.1} & {55.3} & {37.0} \\
    IU-Xray     & R2Gen$^{w\bullet}$ (w/ LDM)                          & {16.6} & {54.6} & {36.0} \\
                & R2Gen$^{w\blacklozenge\bullet}$ (w/ VDM\&LDM)        & {17.3} & {\underline{62.8}} & {36.8} \\
                \cmidrule{2-5}
                & Baseline$^{w\bigstar}$ (w/ \red{RadCARE})                       & {17.0} & {\textbf{63.1}} & {36.3} \\ 
                & Baseline$^{w\bigstar\blacklozenge}$ (w/ \red{RadCARE}\&VDM)     & {17.4}  &  {52.3} & {37.4}  \\
                & Baseline$^{w\bigstar \bullet}$ (w/ \red{RadCARE}\&LDM)          &  {\underline{17.8}} &  {57.3} & {\underline{37.8}}  \\
                & \red{CMCRL}          & {\textbf{19.0}} & {59.2} & {\textbf{39.4}} \\
    \midrule
                & Baseline                                             & {10.1} & {13.0} & {27.0} \\ 
                & Baseline$^{w\blacklozenge}$ (w/ VDM)                 & {10.3} & {14.4} & {27.2}\\
                & Baseline$^{w\bullet}$ (w/ LDM)                       & {6.9} & {7.1} & {22.4}\\
                & Baseline$^{w\blacklozenge \bullet} $ (w/ VDM\&LDM)   & {7.0} & {7.4} & {23.0} \\
                \cmidrule{2-5}
                & R2Gen                                                & {10.3} & {16.8} & {\underline{27.8}}   \\
                & R2Gen$^{w\blacklozenge}$ (w/ VDM)                    & {10.6} & {17.1} & {27.7}  \\
    MIMIC-CXR   & R2Gen$^{w\bullet}$ (w/ LDM)                          & {9.1} & {13.6}  & {25.6}  \\
                & R2Gen$^{w\blacklozenge\bullet}$ (w/ VDM\&LDM)        & {10.0} & {14.3} & {26.4}\\
                \cmidrule{2-5}
                & Baseline$^{w\bigstar}$ (w/ \red{RadCARE})                      & {10.6} & {15.1} & {\underline{27.8}} \\ 
                & Baseline$^{w\bigstar\blacklozenge}$ (w/ \red{RadCARE}\&VDM)    & {11.0} & {\underline{17.7}} & {\textbf{28.0}} \\
                & Baseline$^{w\bigstar \bullet}$ (w/ \red{RadCARE}\&LDM)         & {\underline{11.5}} & {15.7} & {27.7} \\
                & \red{CMCRL}                                                 & {\textbf{11.9}} & {\textbf{19.0}} & {\textbf{28.0}} \\
    \bottomrule
  \end{tabular}}
\end{table}

\subsubsection{Effectiveness of \red{RadCARE}}

In Table~\ref{tab:experiment_vlp_mask}, we conduct ablation experiments to assess the impact of masking ratios on model performance, and the results are presented. Our \red{RadCARE} model achieved the best performance with a higher masking ratio of 85\%, which is in contrast to the optimal masking ratio of 75\% reported by MAE~\cite{he2022masked}. We attribute this difference to the cross-modal information correlations, where the masked information can be reconstructed by visible features from both language and images. Furthermore, \red{RadCARE} tends to learn general features from the masked modality at higher masking ratios, while distinguishable features can be extracted by the complete information from another modality. To explore whether increasing the masking ratio further would further improve the performance, we experimented with a higher masking ratio of 95\%. However, the decreased results in Table~\ref{tab:experiment_vlp_mask} indicate that this approach leads to excessive information loss.

In Table~\ref{tab:ablation_vlp}, we make a comparison with different pre-training methods. It shows that the cross-modal pre-training method has a more robust representation ability than the \red{masked image restoration} with single-modality. However, the Rouge-L metric in VisualGPT surpasses ours, possibly due to its exclusive pre-training of the text decoder, enabling more concentrated learning of the intricate structure in radiological reports.
\red{Additionally, our cross-modal pre-training via postfix text generation task achieves comparable performance to the model that only fine-tunes the encoder, while ours fine-tunes the whole model with fewer epochs. }
Moreover, our \red{RadCARE} adopts the degradation images as input, which facilitates the extraction of visual details in the downstream task.

Furthermore, in Table~\ref{tab:ablation_baseline}, Baseline$^{w\blacklozenge \bullet}$ is significantly worse than baseline on MIMIC-CXR, e.g., 10.1 $\to$ 7.0 for BLEU-4, while still keeping performance improvement on IU-Xray dataset. 
This validates the significant feature complexity from the large-scale MIMIC-CXR dataset leads to unstable probability distribution estimation with causal intervention. 
Meanwhile, the \red{RadCARE} can substantially boost the performance of the baseline, e.g., 14.8 $\to$ 17.0, 10.1 $\to$ 10.6 for BLEU-4 on IU-Xray and MIMIC-CXR datasets, respectively. The improvement is caused by the learned comprehensive concepts and context in the pre-training and the cross-modal features alignment stage, which shows the importance of \red{RadCARE}. Similarly, The Rough-L is also barely improved due to the features' complexity and long sequence from the MIMIC-CXR dataset. For example, although AlignTransformer achieves the same score of the Rough-L as CA on the MIMIC-CXR, it outperforms CA on all other metrics.

\subsubsection{Effectiveness of Causal Intervention}
\noindent \textbf{VDM}. 
In Table~\ref{tab:ablation_baseline}, Baseline$^{w\blacklozenge}$ and R2Gen$^{w\blacklozenge}$ can boost the performance compared to Baseline and R2Gen, which demonstrates the validity of the VDM. 
However, the improvement of BLEU-4 between Baseline$^{w\bigstar \blacklozenge}$ and Baseline$^{w\bigstar}$ on the IU-Xray dataset is more significant than that on the MIMIC-CXR dataset. This is because the VDM can discover essential visual information, but the report of the MIMIC-CXR is more complex and the model fails to generate accurate descriptions. The performance degradation of CIDEr can further illustrate it.

\begin{figure}[t]
 \centering
    \includegraphics[width=1\linewidth]{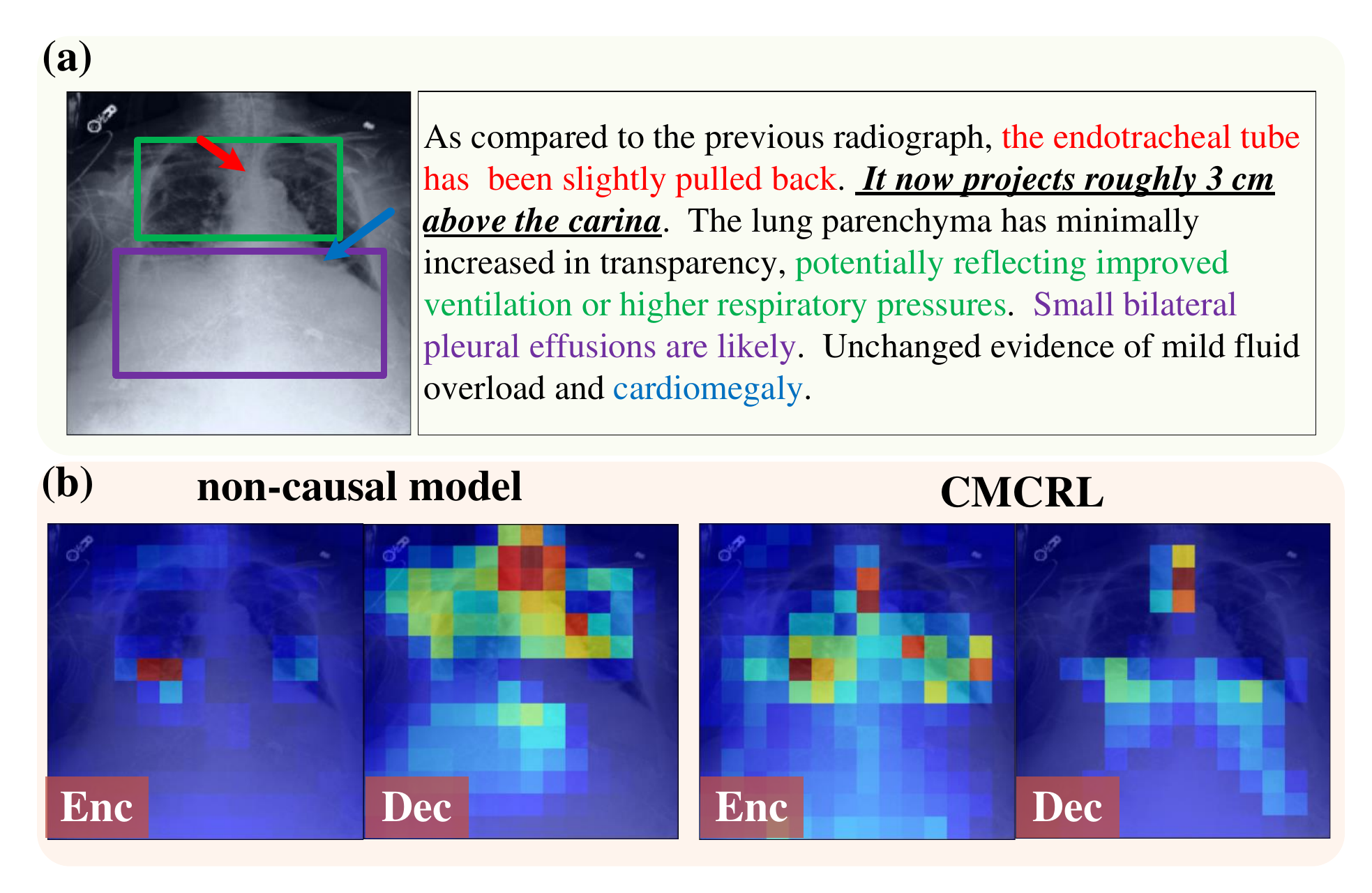}
    \caption{The visualization of the attention map. (a) is an example from the MIMIC-CXR dataset that the colored text should be discovered in the marked region of the image. The images in (b) are the attention maps of the non-causal model and our \red{CMCRL}, respectively. The tag ``Enc" means the accumulated attention maps from the encoder for the selected local features, and ``Dec" is the response to the ``pleural" (decoder output).}
    \label{fig:experiment_local_vis}
\end{figure}

In Fig.~\ref{fig:experiment_local_vis}, the encoder of the non-causal model exhibits limited attention to all potential abnormal regions.  Instead, it excessively focuses on the base of the lung, possibly due to the dataset's high prevalence of lung-related diseases. In contrast, the attention map from the \red{CMCRL} encoder can truly focus on the dominant regions that may indicate abnormalities, including the entire lung lobe, carina, and pleura, rather than false correlations with biased visual concepts. This confirms the semantic sensitivity of the VDM, which captures dominant visual content by performing visual causal interventions.

\noindent \textbf{LDM}. 
Compared to the VDM, the LDM plays a more significant role in {RRG} because the sophisticated linguistic semantic patterns within reports are entangled and biased that require elaborate linguistic deconfounding. 
In Table~\ref{tab:ablation_baseline}, the performance drops without LDM, e.g., 11.9 $\to$ 11.0 for the BLEU-4 metric on the MIMIC-CXR dataset. This shows the importance of adjusting semantic relevance in word embedding space. 
Compared with the baseline, the performance improvement of Baseline$^{w\bigstar \bullet}$ on the MIMIC-CXR dataset demonstrates that the LDM can generate more accurate reports even with biased visual information. 
However, the CIDEr metric on the IU-Xray dataset shows the effectiveness of the combination of VDM and LDM, while ILVD obtains a lower score. This is due to the worse diversity on the IU-Xray dataset, where Baseline$^{w\blacklozenge \bullet}$ and R2Gen$^{w\blacklozenge \bullet}$ can get higher CIDEr but lower BLEU-4 with inadequate multi-modal feature correlation.
In Fig.~\ref{fig:experiment_local_vis} (b), the attention map of the decoder in the non-causal model exhibits evident redundant responses, with high attention widely distributed in the upper part of the lung, especially the carina. 
The \red{CMCRL}, in contrast, can capture dominant semantic information in a coarse-to-fine manner, refining it from the potential abnormal regions that receive extensive attention in the encoder to the bilateral thorax. The high attention on the carina may be attributed to the presence of a support device that could increase cardiac load, cause vascular occlusion or congestion, leading to changes in intrathoracic pressure and eventually resulting in pleural effusion.
These findings indicate that the LDM can capture more discriminative semantic information from the linguistic modality through linguistic front-door interventions.


\section{Conclusion}

\red{In this paper, we \red{propose the Cross-Modal Causal Representation Learning (CMCRL)} framework for {RRG}, to deconfound visual-linguistic confounders by causal intervention. To alleviate the problem of unpaired visual-linguistic data when pre-training, we \red{propose RadCARE} for cross-modal pre-training, \red{integrating postfix text generation and degradation-aware masked image restoration task}. To mitigate cross-modal confounders and discover the true cross-modal causality, we propose visual-linguistic causal front-door intervention modules VDM and LDM\red{, which are integrated into the visual-Linguistic Causal Intervention (VLCI) model}. Experiments on IU-Xray and MIMIC-CXR datasets show that our \red{CMCRL} can effectively mitigate visual-linguistic bias and outperforms the state-of-the-art methods. The lower computational cost and faster inference speed of \red{CMCRL} promote its clinical application.  \red{In future work, we aim to further refine our method to better align with advanced Large Language Models (LLMs) and apply it to a broader range of medical imaging modalities, including CT, MRI, and digital pathology slides, among others. Additionally, to enhance the interpretability of causal reasoning, we will integrate the inferential capabilities of LLMs with their internal knowledge for causal inference. This may involve leveraging existing multimodal data to construct retrieval-augmented generation (RAG) tools or employing multi-agent systems to mine patient causal information from multidimensional medical records. We believe our work will inspire more causal reasoning methods within the realm of Radiology Report Generation (RRG) and extend its impact to other areas of medical imaging.}}

{\small
\bibliography{ref}

\begin{thebibliography}{10}
\providecommand{\url}[1]{#1}
\csname url@samestyle\endcsname
\providecommand{\newblock}{\relax}
\providecommand{\bibinfo}[2]{#2}
\providecommand{\BIBentrySTDinterwordspacing}{\spaceskip=0pt\relax}
\providecommand{\BIBentryALTinterwordstretchfactor}{4}
\providecommand{\BIBentryALTinterwordspacing}{\spaceskip=\fontdimen2\font plus
\BIBentryALTinterwordstretchfactor\fontdimen3\font minus
  \fontdimen4\font\relax}
\providecommand{\BIBforeignlanguage}[2]{{%
\expandafter\ifx\csname l@#1\endcsname\relax
\typeout{** WARNING: IEEEtran.bst: No hyphenation pattern has been}%
\typeout{** loaded for the language `#1'. Using the pattern for}%
\typeout{** the default language instead.}%
\else
\language=\csname l@#1\endcsname
\fi
#2}}
\providecommand{\BIBdecl}{\relax}
\BIBdecl

\bibitem{yu2022crosslink}
Q.~Yu, L.~Qi, Y.~Gao, W.~Wang, and Y.~Shi, ``Crosslink-net: double-branch
  encoder network via fusing vertical and horizontal convolutions for medical
  image segmentation,'' \emph{IEEE Transactions on Image Processing}, vol.~31,
  pp. 5893--5908, 2022.

\bibitem{zhou2021review}
S.~K. Zhou, H.~Greenspan, C.~Davatzikos, J.~S. Duncan, B.~Van~Ginneken,
  A.~Madabhushi, J.~L. Prince, D.~Rueckert, and R.~M. Summers, ``A review of
  deep learning in medical imaging: Imaging traits, technology trends, case
  studies with progress highlights, and future promises,'' \emph{Proceedings of
  the IEEE}, vol. 109, no.~5, pp. 820--838, 2021.

\bibitem{tanida2023interactive}
T.~Tanida, P.~M{\"u}ller, G.~Kaissis, and D.~Rueckert, ``Interactive and
  explainable region-guided radiology report generation,'' in \emph{Proceedings
  of the IEEE/CVF Conference on Computer Vision and Pattern Recognition}, 2023,
  pp. 7433--7442.

\bibitem{nguyen2022effective}
T.-S. Nguyen and B.~Fernando, ``Effective multimodal encoding for image
  paragraph captioning,'' \emph{IEEE Transactions on Image Processing},
  vol.~31, pp. 6381--6395, 2022.

\bibitem{chen2020generating}
Z.~Chen, Y.~Song, T.-H. Chang, and X.~Wan, ``Generating radiology reports via
  memory-driven transformer,'' in \emph{Proceedings of the 2020 Conference on
  Empirical Methods in Natural Language Processing}, 2020, pp. 1439--1449.

\bibitem{zhang2020radiology}
Y.~Zhang, X.~Wang, Z.~Xu, Q.~Yu, A.~Yuille, and D.~Xu, ``When radiology report
  generation meets knowledge graph,'' in \emph{Proceedings of the AAAI
  Conference on Artificial Intelligence}, vol.~34, no.~07, 2020, pp.
  12\,910--12\,917.

\bibitem{chen2023cross}
W.~Chen, Y.~Liu, C.~Wang, J.~Zhu, S.~Zhao, G.~Li, C.-L. Liu, and L.~Lin,
  ``Cross-modal causal intervention for medical report generation,''
  \emph{arXiv preprint arXiv:2303.09117}, 2023.

\bibitem{liu2021exploring}
F.~Liu, X.~Wu, S.~Ge, W.~Fan, and Y.~Zou, ``Exploring and distilling posterior
  and prior knowledge for radiology report generation,'' in \emph{Proceedings
  of the IEEE/CVF Conference on Computer Vision and Pattern Recognition}, 2021,
  pp. 13\,753--13\,762.

\bibitem{liu2021contrastive}
F.~Liu, C.~Yin, X.~Wu, S.~Ge, P.~Zhang, and X.~Sun, ``Contrastive attention for
  automatic chest x-ray report generation,'' in \emph{Findings of the
  Association for Computational Linguistics: ACL-IJCNLP 2021}, 2021, pp.
  269--280.

\bibitem{yang2022knowledge}
S.~Yang, X.~Wu, S.~Ge, S.~K. Zhou, and L.~Xiao, ``Knowledge matters: Chest
  radiology report generation with general and specific knowledge,''
  \emph{Medical Image Analysis}, p. 102510, 2022.

\bibitem{pearl2016causal}
J.~Pearl, M.~Glymour, and N.~P. Jewell, \emph{Causal Inference In Statistics: A
  Primer}.\hskip 1em plus 0.5em minus 0.4em\relax John Wiley \& Sons, 2016.

\bibitem{liu2022causal}
Y.~Liu, Y.-S. Wei, H.~Yan, G.-B. Li, and L.~Lin, ``Causal reasoning meets
  visual representation learning: A prospective study,'' \emph{Machine
  Intelligence Research}, pp. 1--27, 2022.

\bibitem{liu2023cross}
Y.~Liu, G.~Li, and L.~Lin, ``Cross-modal causal relational reasoning for
  event-level visual question answering,'' \emph{IEEE Transactions on Pattern
  Analysis and Machine Intelligence}, 2023.

\bibitem{yang2021causal}
X.~Yang, H.~Zhang, G.~Qi, and J.~Cai, ``Causal attention for vision-language
  tasks,'' in \emph{Proceedings of the IEEE/CVF Conference on Computer Vision
  and Pattern Recognition}, 2021, pp. 9847--9857.

\bibitem{yang2021deconfounded}
X.~Yang, H.~Zhang, and J.~Cai, ``Deconfounded image captioning: A causal
  retrospect,'' \emph{IEEE Transactions on Pattern Analysis and Machine
  Intelligence}, 2021.

\bibitem{liu2023m3ae}
H.~Liu, D.~Wei, D.~Lu, J.~Sun, L.~Wang, and Y.~Zheng, ``M3ae: multimodal
  representation learning for brain tumor segmentation with missing
  modalities,'' in \emph{Proceedings of the AAAI Conference on Artificial
  Intelligence}, vol.~37, no.~2, 2023, pp. 1657--1665.

\bibitem{he2022masked}
K.~He, X.~Chen, S.~Xie, Y.~Li, P.~Doll{\'a}r, and R.~Girshick, ``Masked
  autoencoders are scalable vision learners,'' in \emph{Proceedings of the
  IEEE/CVF Conference on Computer Vision and Pattern Recognition}, 2022, pp.
  16\,000--16\,009.

\bibitem{wang2021simvlm}
Z.~Wang, J.~Yu, A.~W. Yu, Z.~Dai, Y.~Tsvetkov, and Y.~Cao, ``Simvlm: Simple
  visual language model pretraining with weak supervision,'' \emph{arXiv
  preprint arXiv:2108.10904}, 2021.

\bibitem{stefanini2022show}
M.~Stefanini, M.~Cornia, L.~Baraldi, S.~Cascianelli, G.~Fiameni, and
  R.~Cucchiara, ``From show to tell: a survey on deep learning-based image
  captioning,'' \emph{IEEE Transactions on Pattern Analysis and Machine
  Intelligence}, 2022.

\bibitem{karpathy2015deep}
A.~Karpathy and L.~Fei-Fei, ``Deep visual-semantic alignments for generating
  image descriptions,'' in \emph{Proceedings of the IEEE/CVF Conference on
  Computer Vision and Pattern Recognition}, 2015, pp. 3128--3137.

\bibitem{jiang2022visual}
W.~Jiang, M.~Zhu, Y.~Fang, G.~Shi, X.~Zhao, and Y.~Liu, ``Visual cluster
  grounding for image captioning,'' \emph{IEEE Transactions on Image
  Processing}, vol.~31, pp. 3920--3934, 2022.

\bibitem{wang2017skeleton}
Y.~Wang, Z.~Lin, X.~Shen, S.~Cohen, and G.~W. Cottrell, ``Skeleton key: Image
  captioning by skeleton-attribute decomposition,'' in \emph{Proceedings of the
  IEEE/CVF Conference on Computer Vision and Pattern Recognition}, 2017, pp.
  7272--7281.

\bibitem{xian2019self}
Y.~Xian and Y.~Tian, ``Self-guiding multimodal lstm—when we do not have a
  perfect training dataset for image captioning,'' \emph{IEEE Transactions on
  Image Processing}, vol.~28, no.~11, pp. 5241--5252, 2019.

\bibitem{devlin2018bert}
J.~Devlin, M.-W. Chang, K.~Lee, and K.~Toutanova, ``Bert: Pre-training of deep
  bidirectional transformers for language understanding,'' \emph{arXiv preprint
  arXiv:1810.04805}, 2018.

\bibitem{nguyen2022grit}
V.-Q. Nguyen, M.~Suganuma, and T.~Okatani, ``Grit: Faster and better image
  captioning transformer using dual visual features,'' in \emph{European
  Conference on Computer Vision}, 2022, pp. 167--184.

\bibitem{liu2022show}
B.~Liu, D.~Wang, X.~Yang, Y.~Zhou, R.~Yao, Z.~Shao, and J.~Zhao, ``Show,
  deconfound and tell: Image captioning with causal inference,'' in
  \emph{Proceedings of the IEEE/CVF Conference on Computer Vision and Pattern
  Recognition}, 2022, pp. 18\,041--18\,050.

\bibitem{zhou2019re}
L.~Zhou, Y.~Zhang, Y.-G. Jiang, T.~Zhang, and W.~Fan, ``Re-caption:
  Saliency-enhanced image captioning through two-phase learning,'' \emph{IEEE
  Transactions on Image Processing}, vol.~29, pp. 694--709, 2019.

\bibitem{yu2022coca}
J.~Yu, Z.~Wang, V.~Vasudevan, L.~Yeung, M.~Seyedhosseini, and Y.~Wu, ``Coca:
  Contrastive captioners are image-text foundation models,'' \emph{arXiv
  preprint arXiv:2205.01917}, 2022.

\bibitem{zhu2024beyond}
L.~Zhu, F.~Wei, and Y.~Lu, ``Beyond text: Frozen large language models in
  visual signal comprehension,'' in \emph{Proceedings of the IEEE/CVF
  Conference on Computer Vision and Pattern Recognition}, 2024, pp.
  27\,047--27\,057.

\bibitem{liu2024improved}
H.~Liu, C.~Li, Y.~Li, and Y.~J. Lee, ``Improved baselines with visual
  instruction tuning,'' in \emph{Proceedings of the IEEE/CVF Conference on
  Computer Vision and Pattern Recognition}, 2024, pp. 26\,296--26\,306.

\bibitem{chen2024sharegpt4v}
L.~Chen, J.~Li, X.~Dong, P.~Zhang, C.~He, J.~Wang, F.~Zhao, and D.~Lin,
  ``Sharegpt4v: Improving large multi-modal models with better captions,'' in
  \emph{European Conference on Computer Vision}.\hskip 1em plus 0.5em minus
  0.4em\relax Springer, 2024, pp. 370--387.

\bibitem{hurst2024gpt}
A.~Hurst, A.~Lerer, A.~P. Goucher, A.~Perelman, A.~Ramesh, A.~Clark, A.~Ostrow,
  A.~Welihinda, A.~Hayes, A.~Radford \emph{et~al.}, ``Gpt-4o system card,''
  \emph{arXiv preprint arXiv:2410.21276}, 2024.

\bibitem{bai2023qwen}
J.~Bai, S.~Bai, Y.~Chu, Z.~Cui, K.~Dang, X.~Deng, Y.~Fan, W.~Ge, Y.~Han,
  F.~Huang \emph{et~al.}, ``Qwen technical report,'' \emph{arXiv preprint
  arXiv:2309.16609}, 2023.

\bibitem{yue2020interventional}
Z.~Yue, H.~Zhang, Q.~Sun, and X.-S. Hua, ``Interventional few-shot learning,''
  \emph{Advances in Neural Information Processing Systems}, vol.~33, 2020.

\bibitem{miao2023caussl}
J.~Miao, C.~Chen, F.~Liu, H.~Wei, and P.-A. Heng, ``Caussl: Causality-inspired
  semi-supervised learning for medical image segmentation,'' in
  \emph{Proceedings of the IEEE/CVF International Conference on Computer
  Vision}, 2023, pp. 21\,426--21\,437.

\bibitem{liu2022contextual}
R.~Liu, H.~Liu, G.~Li, H.~Hou, T.~Yu, and T.~Yang, ``Contextual debiasing for
  visual recognition with causal mechanisms,'' in \emph{Proceedings of the
  IEEE/CVF Conference on Computer Vision and Pattern Recognition}, 2022, pp.
  12\,755--12\,765.

\bibitem{wang2021causal}
T.~Wang, C.~Zhou, Q.~Sun, and H.~Zhang, ``Causal attention for unbiased visual
  recognition,'' in \emph{Proceedings of the IEEE/CVF International Conference
  on Computer Vision}, 2021, pp. 3091--3100.

\bibitem{wang2020visual}
T.~Wang, J.~Huang, H.~Zhang, and Q.~Sun, ``Visual commonsense representation
  learning via causal inference,'' in \emph{Proceedings of the IEEE/CVF
  Conference on Computer Vision and Pattern Recognition Workshops}, 2020, pp.
  378--379.

\bibitem{zang2023discovering}
C.~Zang, H.~Wang, M.~Pei, and W.~Liang, ``Discovering the real association:
  Multimodal causal reasoning in video question answering,'' in
  \emph{Proceedings of the IEEE/CVF Conference on Computer Vision and Pattern
  Recognition}, 2023, pp. 19\,027--19\,036.

\bibitem{yang2023good}
Z.~Yang, M.~Lin, X.~Zhong, Y.~Wu, and Z.~Wang, ``Good is bad: Causality
  inspired cloth-debiasing for cloth-changing person re-identification,'' in
  \emph{Proceedings of the IEEE/CVF Conference on Computer Vision and Pattern
  Recognition}, 2023, pp. 1472--1481.

\bibitem{hu2021causal}
Z.~Hu and L.~E. Li, ``A causal lens for controllable text generation,''
  \emph{Advances in Neural Information Processing Systems}, vol.~34, pp.
  24\,941--24\,955, 2021.

\bibitem{xue2023variational}
D.~Xue, S.~Qian, and C.~Xu, ``Variational causal inference network for
  explanatory visual question answering,'' in \emph{Proceedings of the IEEE/CVF
  International Conference on Computer Vision}, 2023, pp. 2515--2525.

\bibitem{zhao2024causal}
S.~Zhao, Z.~Li, Y.~Lu, A.~Yuille, and Y.~Wang, ``Causal-cog: A causal-effect
  look at context generation for boosting multi-modal language models,'' in
  \emph{Proceedings of the IEEE/CVF Conference on Computer Vision and Pattern
  Recognition}, 2024, pp. 13\,342--13\,351.

\bibitem{chu2024causal}
Z.~Chu, Y.~Wang, L.~Li, Z.~Wang, Z.~Qin, and K.~Ren, ``A causal explainable
  guardrails for large language models,'' in \emph{Proceedings of the 2024 on
  ACM SIGSAC Conference on Computer and Communications Security}, 2024, pp.
  1136--1150.

\bibitem{huang2025causality}
Z.~Huang, S.~Zhong, P.~Zhou, S.~Gao, M.~Zitnik, and L.~Lin, ``A causality-aware
  paradigm for evaluating creativity of multimodal large language models,''
  \emph{IEEE Transactions on Pattern Analysis and Machine Intelligence}, pp.
  1--17, 2025.

\bibitem{zhao2018causaltriad}
S.~Zhao, M.~Jiang, M.~Liu, B.~Qin, and T.~Liu, ``Causaltriad: toward pseudo
  causal relation discovery and hypotheses generation from medical text data,''
  in \emph{Proceedings of the 2018 ACM International Conference on
  Bioinformatics, Computational Biology, and Health Informatics}, 2018, pp.
  184--193.

\bibitem{li2023causally}
X.~Li, X.~Qian, L.~Liang, L.~Kong, Q.~Dong, J.~Chen, D.~Liu, X.~Yao, and Y.~Fu,
  ``Causally-aware intraoperative imputation for overall survival time
  prediction,'' in \emph{Proceedings of the IEEE/CVF Conference on Computer
  Vision and Pattern Recognition}, 2023, pp. 15\,681--15\,690.

\bibitem{jain2021radgraph}
S.~Jain, A.~Agrawal, A.~Saporta, S.~Q. Truong, D.~N. Duong, T.~Bui, P.~Chambon,
  Y.~Zhang, M.~P. Lungren, A.~Y. Ng \emph{et~al.}, ``Radgraph: Extracting
  clinical entities and relations from radiology reports,'' \emph{arXiv
  preprint arXiv:2106.14463}, 2021.

\bibitem{li2023dynamic}
M.~Li, B.~Lin, Z.~Chen, H.~Lin, X.~Liang, and X.~Chang, ``Dynamic graph
  enhanced contrastive learning for chest x-ray report generation,'' in
  \emph{Proceedings of the IEEE/CVF Conference on Computer Vision and Pattern
  Recognition}, 2023, pp. 3334--3343.

\bibitem{you2021aligntransformer}
D.~You, F.~Liu, S.~Ge, X.~Xie, J.~Zhang, and X.~Wu, ``Aligntransformer:
  Hierarchical alignment of visual regions and disease tags for medical report
  generation,'' in \emph{International Conference on Medical Image Computing
  and Computer-Assisted Intervention}.\hskip 1em plus 0.5em minus 0.4em\relax
  Springer, 2021, pp. 72--82.

\bibitem{wang2022cross}
J.~Wang, A.~Bhalerao, and Y.~He, ``Cross-modal prototype driven network for
  radiology report generation,'' in \emph{European Conference on Computer
  Vision}.\hskip 1em plus 0.5em minus 0.4em\relax Springer, 2022, pp. 563--579.

\bibitem{liu2022competence}
F.~Liu, S.~Ge, and X.~Wu, ``Competence-based multimodal curriculum learning for
  medical report generation,'' \emph{arXiv preprint arXiv:2206.14579}, 2022.

\bibitem{wang2024camanet}
J.~Wang, A.~Bhalerao, T.~Yin, S.~See, and Y.~He, ``Camanet: class activation
  map guided attention network for radiology report generation,'' \emph{IEEE
  Journal of Biomedical and Health Informatics}, vol.~28, no.~4, pp.
  2199--2210, 2024.

\bibitem{nooralahzadeh2021progressive}
F.~Nooralahzadeh, N.~P. Gonzalez, T.~Frauenfelder, K.~Fujimoto, and
  M.~Krauthammer, ``Progressive transformer-based generation of radiology
  reports,'' \emph{arXiv preprint arXiv:2102.09777}, 2021.

\bibitem{divya2024memory}
P.~Divya, Y.~Sravani, C.~Vishnu, C.~K. Mohan, and Y.~W. Chen, ``Memory guided
  transformer with spatio-semantic visual extractor for medical report
  generation,'' \emph{IEEE Journal of Biomedical and Health Informatics}, 2024.

\bibitem{chen2022cross}
Z.~Chen, Y.~Shen, Y.~Song, and X.~Wan, ``Cross-modal memory networks for
  radiology report generation,'' \emph{arXiv preprint arXiv:2204.13258}, 2022.

\bibitem{liu2024context}
R.~Liu, M.~Li, S.~Zhao, L.~Chen, X.~Chang, and L.~Yao, ``In-context learning
  for zero-shot medical report generation,'' in \emph{Proceedings of the 32nd
  ACM International Conference on Multimedia}, 2024, pp. 8721--8730.

\bibitem{jin2024promptmrg}
H.~Jin, H.~Che, Y.~Lin, and H.~Chen, ``Promptmrg: Diagnosis-driven prompts for
  medical report generation,'' in \emph{Proceedings of the AAAI Conference on
  Artificial Intelligence}, vol.~38, no.~3, 2024, pp. 2607--2615.

\bibitem{wang2023r2gengpt}
Z.~Wang, L.~Liu, L.~Wang, and L.~Zhou, ``R2gengpt: Radiology report generation
  with frozen llms,'' \emph{Meta-Radiology}, vol.~1, no.~3, p. 100033, 2023.

\bibitem{zhao2023mamo}
Z.~Zhao, L.~Guo, X.~He, S.~Shao, Z.~Yuan, and J.~Liu, ``Mamo: Fine-grained
  vision-language representations learning with masked multimodal modeling,''
  in \emph{Proceedings of the 46th International ACM SIGIR Conference on
  Research and Development in Information Retrieval}, 2023, pp. 1528--1538.

\bibitem{xvlm}
Y.~Zeng, X.~Zhang, and H.~Li, ``Multi-grained vision language pre-training:
  Aligning texts with visual concepts,'' \emph{International Conference on
  Machine Learning}, pp. 25\,994--26\,009, 2022.

\bibitem{singh2022flava}
A.~Singh, R.~Hu, V.~Goswami, G.~Couairon, W.~Galuba, M.~Rohrbach, and D.~Kiela,
  ``Flava: A foundational language and vision alignment model,'' in
  \emph{Proceedings of the IEEE/CVF Conference on Computer Vision and Pattern
  Recognition}, 2022, pp. 15\,638--15\,650.

\bibitem{varma2023villa}
M.~Varma, J.-B. Delbrouck, S.~Hooper, A.~Chaudhari, and C.~Langlotz, ``Villa:
  Fine-grained vision-language representation learning from real-world data,''
  in \emph{Proceedings of the IEEE/CVF International Conference on Computer
  Vision}, 2023, pp. 22\,225--22\,235.

\bibitem{wang2022image}
W.~Wang, H.~Bao, L.~Dong, J.~Bjorck, Z.~Peng, Q.~Liu, K.~Aggarwal, O.~K.
  Mohammed, S.~Singhal, S.~Som \emph{et~al.}, ``Image as a foreign language:
  Beit pretraining for all vision and vision-language tasks,'' \emph{arXiv
  preprint arXiv:2208.10442}, 2022.

\bibitem{tanno2019learning}
R.~Tanno, A.~Saeedi, S.~Sankaranarayanan, D.~C. Alexander, and N.~Silberman,
  ``Learning from noisy labels by regularized estimation of annotator
  confusion,'' in \emph{Proceedings of the IEEE/CVF conference on computer
  vision and pattern recognition}, 2019, pp. 11\,244--11\,253.

\bibitem{wang2024robust}
H.~Wang, J.~He, H.~Cui, B.~Yuan, and Y.~Xia, ``Robust stochastic neural
  ensemble learning with noisy labels for thoracic disease classification,''
  \emph{IEEE Transactions on Medical Imaging}, 2024.

\bibitem{he2016deep}
K.~He, X.~Zhang, S.~Ren, and J.~Sun, ``Deep residual learning for image
  recognition,'' in \emph{Proceedings of the IEEE/CVF Conference on Computer
  Vision and Pattern Recognition}, 2016, pp. 770--778.

\bibitem{geng2022multimodal}
X.~Geng, H.~Liu, L.~Lee, D.~Schuurams, S.~Levine, and P.~Abbeel, ``Multimodal
  masked autoencoders learn transferable representations,'' \emph{arXiv
  preprint arXiv:2205.14204}, 2022.

\bibitem{zhou2023advancing}
H.-Y. Zhou, C.~Lian, L.~Wang, and Y.~Yu, ``Advancing radiograph representation
  learning with masked record modeling,'' \emph{arXiv preprint
  arXiv:2301.13155}, 2023.

\bibitem{huang2021gloria}
S.-C. Huang, L.~Shen, M.~P. Lungren, and S.~Yeung, ``Gloria: A multimodal
  global-local representation learning framework for label-efficient medical
  image recognition,'' in \emph{Proceedings of the IEEE/CVF International
  Conference on Computer Vision}, 2021, pp. 3942--3951.

\bibitem{cheng2023prior}
P.~Cheng, L.~Lin, J.~Lyu, Y.~Huang, W.~Luo, and X.~Tang, ``Prior: Prototype
  representation joint learning from medical images and reports,'' in
  \emph{Proceedings of the IEEE/CVF International Conference on Computer
  Vision}, 2023, pp. 21\,361--21\,371.

\bibitem{lopez2017discovering}
D.~Lopez-Paz, R.~Nishihara, S.~Chintala, B.~Scholkopf, and L.~Bottou,
  ``Discovering causal signals in images,'' in \emph{Proceedings of the
  IEEE/CVF Conference on Computer Vision and Pattern Recognition}, 2017, pp.
  6979--6987.

\bibitem{qi2020two}
J.~Qi, Y.~Niu, J.~Huang, and H.~Zhang, ``Two causal principles for improving
  visual dialog,'' in \emph{Proceedings of the IEEE/CVF Conference on Computer
  Vision and Pattern Recognition}, 2020, pp. 10\,860--10\,869.

\bibitem{xu2015show}
K.~Xu, J.~Ba, R.~Kiros, K.~Cho, A.~Courville, R.~Salakhudinov, R.~Zemel, and
  Y.~Bengio, ``Show, attend and tell: Neural image caption generation with
  visual attention,'' in \emph{ICML}.\hskip 1em plus 0.5em minus 0.4em\relax
  PMLR, 2015, pp. 2048--2057.

\bibitem{sun2022lesion}
J.~Sun, D.~Wei, L.~Wang, and Y.~Zheng, ``Lesion guided explainable few
  weak-shot medical report generation,'' in \emph{International Conference on
  Medical Image Computing and Computer-Assisted Intervention}.\hskip 1em plus
  0.5em minus 0.4em\relax Springer, 2022, pp. 615--625.

\bibitem{he2022transfg}
J.~He, J.-N. Chen, S.~Liu, A.~Kortylewski, C.~Yang, Y.~Bai, and C.~Wang,
  ``Transfg: A transformer architecture for fine-grained recognition,'' in
  \emph{International Conference on Medical Image Computing and
  Computer-Assisted Intervention}, vol.~36, no.~1, 2022, pp. 852--860.

\bibitem{dai2021coatnet}
Z.~Dai, H.~Liu, Q.~V. Le, and M.~Tan, ``Coatnet: Marrying convolution and
  attention for all data sizes,'' \emph{Advances in Neural Information
  Processing Systems}, vol.~34, pp. 3965--3977, 2021.

\bibitem{li2020sentence}
B.~Li, H.~Zhou, J.~He, M.~Wang, Y.~Yang, and L.~Li, ``On the sentence
  embeddings from pre-trained language models,'' \emph{arXiv preprint
  arXiv:2011.05864}, 2020.

\bibitem{cao2023mmtn}
Y.~Cao, L.~Cui, L.~Zhang, F.~Yu, Z.~Li, and Y.~Xu, ``Mmtn: Multi-modal memory
  transformer network for image-report consistent medical report generation,''
  in \emph{Proceedings of the AAAI Conference on Artificial Intelligence},
  vol.~37, no.~1, 2023, pp. 277--285.

\bibitem{zhang2023semi}
K.~Zhang, H.~Jiang, J.~Zhang, Q.~Huang, J.~Fan, J.~Yu, and W.~Han,
  ``Semi-supervised medical report generation via graph-guided hybrid feature
  consistency,'' \emph{IEEE Transactions on Multimedia}, 2023.

\bibitem{jamiaocv080}
D.~Demner-Fushman, M.~D. Kohli, M.~B. Rosenman, S.~E. Shooshan, L.~Rodriguez,
  S.~Antani, G.~R. Thoma, and C.~J. McDonald, ``Preparing a collection of
  radiology examinations for distribution and retrieval,'' \emph{Journal of the
  American Medical Informatics Association}, vol.~23, no.~2, pp. 304--310,
  2015.

\bibitem{johnson2019mimic}
A.~E. Johnson, T.~J. Pollard, N.~R. Greenbaum, M.~P. Lungren, C.-y. Deng,
  Y.~Peng, Z.~Lu, R.~G. Mark, S.~J. Berkowitz, and S.~Horng, ``Mimic-cxr-jpg, a
  large publicly available database of labeled chest radiographs,'' \emph{arXiv
  preprint arXiv:1901.07042}, 2019.

\bibitem{papineni2002bleu}
K.~Papineni, S.~Roukos, T.~Ward, and W.-J. Zhu, ``Bleu: a method for automatic
  evaluation of machine translation,'' in \emph{Proceedings of the 40th annual
  meeting of the Association for Computational Linguistics}, 2002, pp.
  311--318.

\bibitem{rouge2004package}
L.~C. ROUGE, ``A package for automatic evaluation of summaries,'' in
  \emph{Proceedings of Workshop on Text Summarization of ACL}, 2004.

\bibitem{banerjee2005meteor}
S.~Banerjee and A.~Lavie, ``Meteor: An automatic metric for mt evaluation with
  improved correlation with human judgments,'' in \emph{Proceedings of the ACL
  Workshop on Intrinsic and Extrinsic Evaluation Measures for Machine
  Translation and/or Summarization}, 2005, pp. 65--72.

\bibitem{vedantam2015cider}
R.~Vedantam, C.~Lawrence~Zitnick, and D.~Parikh, ``Cider: Consensus-based image
  description evaluation,'' in \emph{Proceedings of the IEEE/CVF Conference on
  Computer Vision and Pattern Recognition}, 2015, pp. 4566--4575.

\bibitem{irvin2019chexpert}
J.~Irvin, P.~Rajpurkar, M.~Ko, Y.~Yu, S.~Ciurea-Ilcus, C.~Chute, H.~Marklund,
  B.~Haghgoo, R.~Ball, K.~Shpanskaya \emph{et~al.}, ``Chexpert: A large chest
  radiograph dataset with uncertainty labels and expert comparison,'' in
  \emph{International Conference on Medical Image Computing and
  Computer-Assisted Intervention}, vol.~33, no.~01, 2019, pp. 590--597.

\bibitem{Chen_2022_CVPR}
J.~Chen, H.~Guo, K.~Yi, B.~Li, and M.~Elhoseiny, ``Visualgpt: Data-efficient
  adaptation of pretrained language models for image captioning,'' in
  \emph{Proceedings of the IEEE/CVF Conference on Computer Vision and Pattern
  Recognition}, June 2022, pp. 18\,030--18\,040.

\end{thebibliography}
\bibliographystyle{IEEEtran}
}

\begin{IEEEbiography}[{\includegraphics[width=1in,height=1.25in,clip,keepaspectratio]{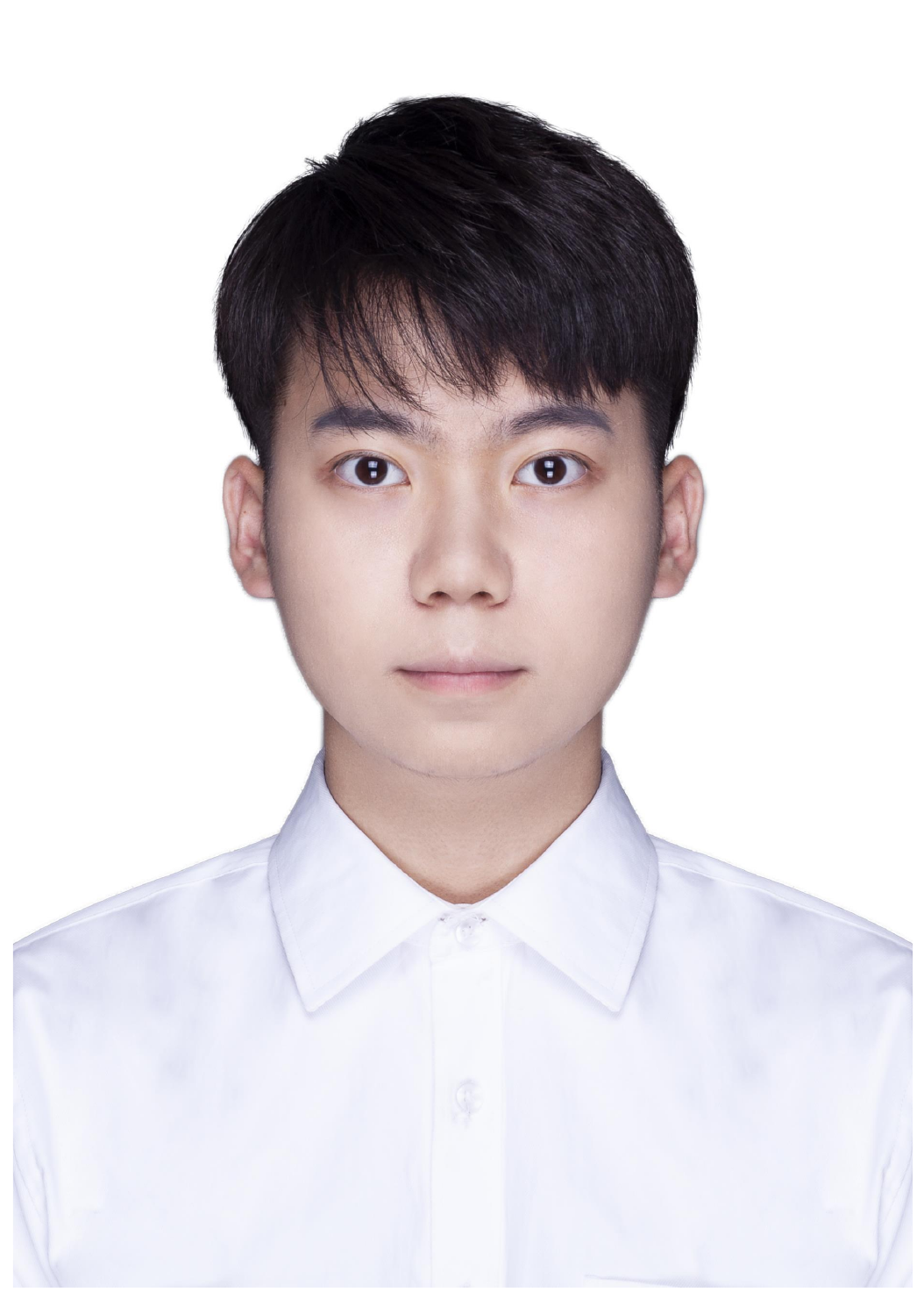}}]{Weixing Chen} has received the B.S. degree from the college of Medicine and Biological Information Engineering, Northeastern University, in 2020 and M.S. degree from Shenzhen Institute of Advanced Technology, Chinese Academy of Sciences in 2023. He is currently a Ph.D. student at the School of Computer Science and Engineering, Sun Yat-sen University. His main interests include medical image analysis, multi-modal learning, and causal relation discovery. He has been serving as a reviewer for numerous academic journals and conferences such as TNNLS, NeurIPS, ICML, ICLR, MICCAI, and ACM MM.
\end{IEEEbiography}

\begin{IEEEbiography}[{\includegraphics[width=1in,height=1.25in,clip,keepaspectratio]{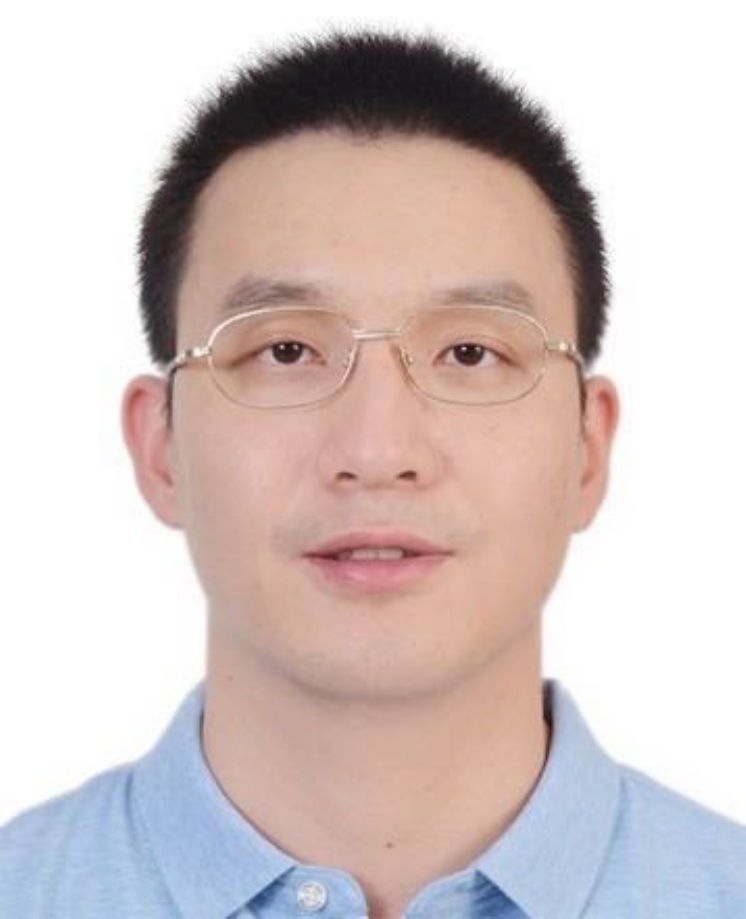}}]{Yang Liu}(M'21) is currently an associate professor working at the School of Computer Science and Engineering, Sun Yat-sen University. He received his Ph.D. degree from Xidian University in 2019. His current research interests include multi-modal reasoning, causality learning and embodied AI. He is the recipient of the First Prize of the Third Guangdong Province Young Computer Science Academic Show. He has authorized and co-authorized more than 40 papers in top-tier academic journals and conferences such as TPAMI, TIP, TKDE, T-MECH, CVPR, ICCV, IJCAI, and ACM MM.
\end{IEEEbiography}

\begin{IEEEbiography}[{\includegraphics[width=1in,height=1.25in,clip,keepaspectratio]{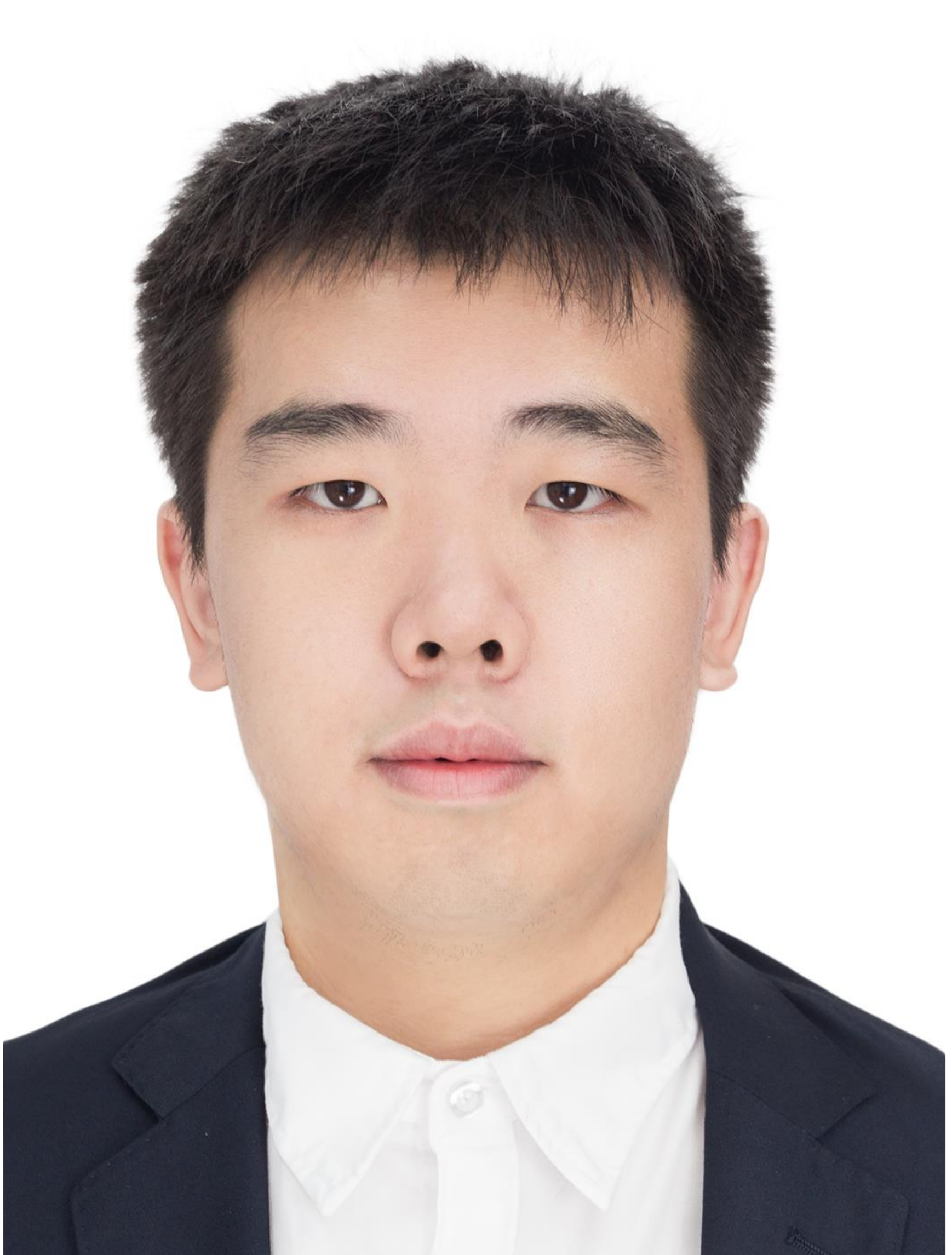}}]{Ce Wang} has received the B.S. degree from the Department of Mathematics, Jilin University, in 2015 and Ph.D. degree from the Center for Combinatorics, Nankai University in 2020. From 2020 to 2023, he worked as a postdoctoral researcher in the Institute of Computing Technology, Chinese Academy of Sciences. From 2023 to 2024, He worked as a postdoctoral researcher in the department of ECE, HKUST. He is currently an Associate Professor with the Sun Yat-sen University, Shenzhen, Guangdong, China. His main interests include image and video processing, computational imaging, medical image analysis, and explainable healthcare AI. He has been serving as a reviewer for numerous academic journals and conferences such as TMI, MIA, TCSVT, NIPS, ICLR, MICCAI and ACM MM.
\end{IEEEbiography}

\begin{IEEEbiography}[{\includegraphics[width=1in,height=1.25in,clip,keepaspectratio]{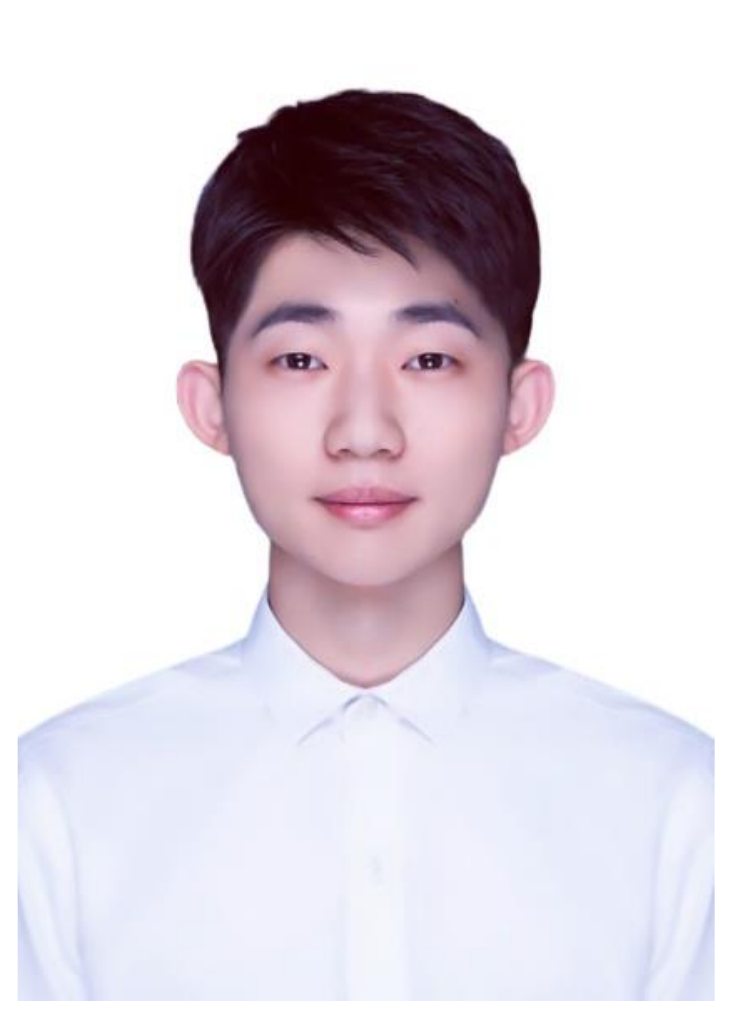}}]{Jiarui Zhu} is currently a Ph.D student in Department of Health Technology and Informatics, the Hong Kong Polytechnic University. He received his B.S. degree from the college of Medicine and Biological Information Engineering, Northeastern University in 2020 and Msc degree in Medical Physics from Hong Kong Polytechnic University in 2023. His current research interests include 3D gaussian representations and few-shot learning in medical imaging processing.
\end{IEEEbiography}

\begin{IEEEbiography}[{\includegraphics[width=1in,height=1.25in,clip,keepaspectratio]{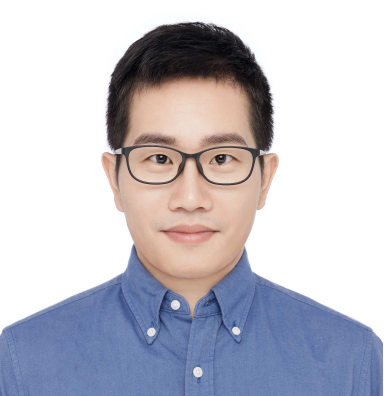}}]{Guanbin Li}(M'15) is currently a professor in School of Computer Science and Engineering, Sun Yat-Sen University. He received his PhD degree from the University of Hong Kong in 2016. His current research interests include computer vision, image processing, and deep learning. He is a recipient of ICCV 2019 Best Paper Nomination Award. He has authorized and co-authorized on more than 100 papers in top-tier academic journals and conferences. He serves as an area chair for the conference of VISAPP. He has been serving as a reviewer for numerous academic journals and conferences such as TPAMI, IJCV, TIP, TMM, TCyb, CVPR, ICCV, ECCV and NeurIPS.
\end{IEEEbiography}

\begin{IEEEbiography}[{\includegraphics[width=1in,height=1.25in,clip,keepaspectratio]{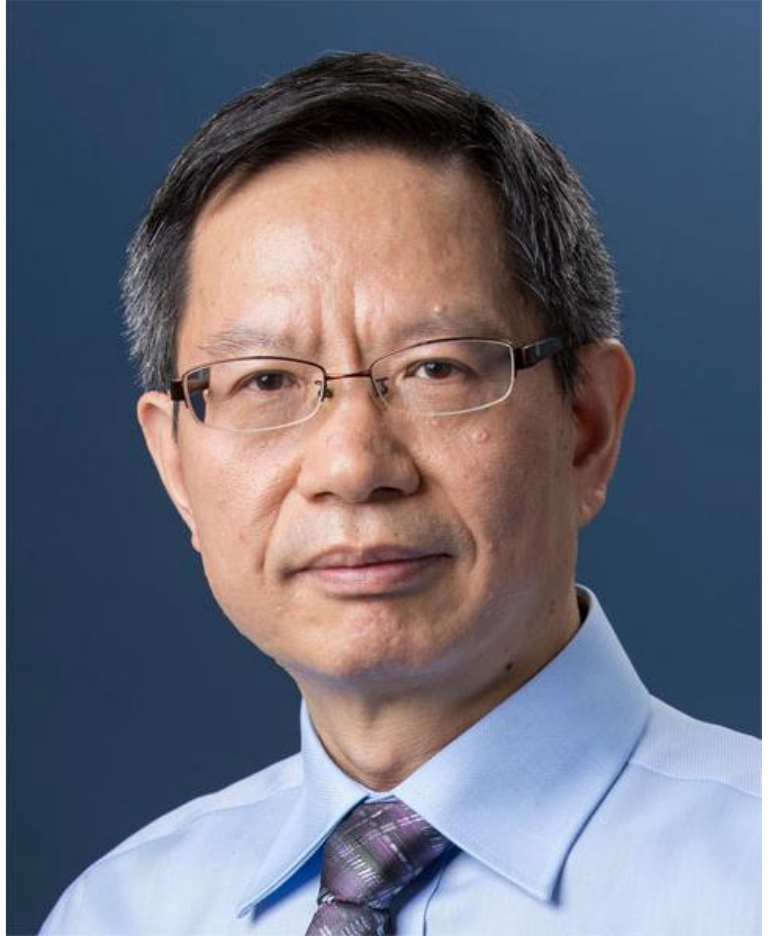}}]{Cheng-Lin Liu}(Fellow, IEEE) received the BS, ME, and PhD degrees from Wuhan University, Beijing University of Technology, and the Institute of Automation of Chinese Academy of Sciences, in 1989,
1992 and 1995, respectively. He was a postdoctoral fellow with the Korea Advanced Institute of Science and Technology (KAIST) and later with the Tokyo University of Agriculture and Technology from 1996 to 1999. From 1999 to 2004, he was a research staff member and later a senior researcher with the Central Research Laboratory, Hitachi, Ltd., Tokyo, Japan.
Since 2005, he has been a professor with the National Laboratory of Pattern Recognition (NLPR), Institute of Automation of Chinese Academy of Sciences, Beijing, China. His research interests include pattern recognition, machine learning, document analysis and recognition. He has published more than 400
technical papers in prestigious international journals and conferences. He is an associate editor-in-chief of Pattern Recognition Journal and Acta Automatica
Sinica and is on the editorial board of several international and domestic journals. He is a fellow of the IAPR, the CAA, and CAAI.
\end{IEEEbiography}

\begin{IEEEbiography}[{\includegraphics[width=1in,height=1.25in,clip,keepaspectratio]{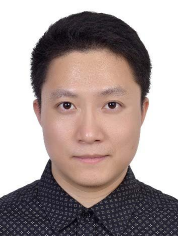}}]{Liang Lin}(Fellow, IEEE) is currently a Full Professor with Sun Yat-sen University, Guangzhou, China. From 2008 to 2010, he was a Postdoctoral Fellow
with the University of California, Los Angeles, Los
Angeles, CA, USA. From 2016 to 2018, he led the SenseTime R\&D teams to develop cutting-edge and deliverable solutions for computer vision, data analysis and mining, and intelligent robotic systems. He
has authored or coauthored more than 100 papers
in top-tier academic journals and conferences, such
as 15 papers in IEEE TRANSACTIONS ON PATTERN
ANALYSIS AND MACHINE I NTELLIGENCE and International Journal of Computer Vision, and more than 60 papers in CVPR, ICCV, NeurIPS, and IJCAI.
He was an Associate Editor for IEEE TRANSACTIONS ON MULTIMEDIA , IEEE TRANSACTIONS ON NEURAL NETWORKS AND LEARNING SYSTEMS, and was an Area/Session Chair for numerous conferences, such as CVPR, ICCV, AAAI,
ICME, and ICMR. He was the recipient of the Annual Best Paper Award by Pattern Recognition (Elsevier) in 2018, Best Paper Diamond Award at IEEE ICME 2017, Best Paper Runner-Up Award at ACM NPAR 2010, Google Faculty Award in 2012, Best Student Paper Award at IEEE ICME 2014, and Hong
Kong Scholars Award in 2014. He is a Fellow of IAPR, AAIA, and IET.
\end{IEEEbiography}


\end{document}